\documentclass[12pt]{article}
\usepackage{graphicx}
\usepackage{amsmath}
\usepackage{epsfig}
\usepackage{amssymb}
\usepackage{hyperref}
\usepackage[authoryear,round]{natbib} 
\newtheorem{assump}{Assumption}

\newtheorem{claim}{Claim}

\def\bSigma{\mbox{\boldmath $\Sigma$}}

\def\mR{\mathbb{R}}

\def\bmu{\mbox{\boldmath $\mu$}}
\def\bepsilon{\mbox{\boldmath $\epsilon$}}

\def\bz{{\bf z}}

\def\bI{{\bf I}}

\def\bQ{{\bf Q}}
\def\vR{{\bf R}}

\def\bX{{\bf X}}

\def\bw{{\bf w}}

\def\bu{{\bf u}}
\def\bx{{\bf x}}
\def\bv{{\bf v}}
\def\by{{\bf y}}
\def\nn{\nonumber}

\def\v2{\vspace{0.2in}}

\begin{document}
\setcounter{page}{1}
\baselineskip=17pt
\footskip=.3in

\title{Large dimensional analysis of general margin based classification methods}

\author{Hanwen Huang \\\\
    {\it Department of Epidemiology and Biostatistics}\\
    {\it University of Georgia}\\
    { Athens, GA 30602, USA}\\
    huanghw@uga.edu\\\\
     Qinglong Yang\footnote{Corresponding author} \\\\
       {\it School of Statistics and Mathematics}\\
       {\it Zhongnan University of Economics and Law}\\
       {\it Wuhan, Hubei 430073, P. R. China}\\
      yangqinglong@zuel.edu.cn}
\date{}

\maketitle

\begin{abstract}
Margin-based classifiers have been popular in both machine learning and statistics for classification problems. Since a large number of classifiers are available, one natural question is which type of classifiers should be used given a particular classification task. We answer this question by investigating the asymptotic performance of a family of large-margin classifiers under the two component mixture models in situations where the data dimension $p$ and the sample $n$ are both large. This family covers a broad range of classifiers including support vector machine, distance weighted discrimination, penalized logistic regression, and large-margin unified machine as special cases. The asymptotic results are described by a set of nonlinear equations and we observe a close match of them with Monte Carlo simulation on finite data samples. Our analytical studies shed new light on how to select the best classifier among various classification methods as well as on  how to choose the optimal tuning parameters for a given method. 
\end{abstract}

\noindent%
  SVM, DWD, logistic regression, nonlinear equation, tuning parameter
\vfill

\section{Introduction}
Classification is a very useful statistical tool which has been widely used in many disciplines and has achieved a lot of success. Its goal is to build a classification rule
based on a training set which includes both covariates and class labels.  Then for new objects whose covariates are available the classification rule can be used for class label prediction.

Since a large number of classifiers are available on the shelf, one natural question to ask is which type of classifiers should be used given a particular classification task. It is commonly agreed upon that there is no single method working best for all problems. The choice of classifiers really depends on the nature of the data set and the primary learning goal. Cross validation (CV) is a practically useful strategy for handling this task; its basic concept is to evaluate the prediction error by examining the data under control. Smaller values of the CV error are expected to be better in expressing the generative model of the data. However, the implementation of many classification methods involves  tuning open parameters for achieving optimal performances, e.g. for regularized classification methods, one needs to deal with tuning parameters that control the trade-off between data fitting and principle of parsimony. Therefore, conducting CV incurs high computational costs, which makes it difficult in practice.

The purpose of this paper is to answer the above question by investigating the asymptotic performance of a family of large-margin classifiers in the limit of both sample size $n$ and dimension $p$ going to infinity with fixed rate $\alpha=n/p$. We are motivated by the comparison between two commonly used classification methods: support vector machine (SVM) and distance weighted discrimination (DWD). 

SVM is a state-of-the-art powerful classification method proposed by Vapnik \citep{Vapnik95}.  Its has been demonstrated in \cite{JMLR:v15:delgado14a} as one of the best performers in the pool of 179 commonly used classifiers. However, as pointed out by \cite{Marron2007}, SVM may suffer from a loss of generalization ability in the high-dimension-low-sample size (HDLSS) setting (where $n$ is much less than $p$) due to data-piling problem. They proposed DWD as a superior alternative to SVM. Both SVM and DWD are margin-based classification methods in the sense that they build the classifier through finding a decision boundary to separate the classes.  DWD is different from SVM in that it seeks to maximize a notion of average distance instead of minimum distance between the classes. Thus, DWD allows all data points, rather than only the support vectors, to have a direct impact on the separating hyperplane. It gives high significance to those points that are close to the hyperplane, with little impact from points that are farther away. DWD is specifically designed for HDLSS situations. Many previous simulations and real data studies have shown that DWD performs better than SVM especially in HDLSS cases, see e.g. \cite{Benito04,Qiao2011,qiao15,zou2,zou1}. However, all previous studies are empirical and there is no theoretical justification about this phenomenon yet.   

%\cite{Hall2008} studied the HDLSS asymptotics of SVM and DWD and showed that for fixed $n$,  as $p\rightarrow\infty$ the classification performance depends on the signal size $\mu$ which is defined as the distance between the mean positions of two classes. Assume that $\mu$ increases with $p$ as $p^\gamma$, then if $\gamma\textgreater 1/2$, both SVM and DWD are strongly consistent, i.e., they can make perfect separation; if $\gamma\textless 1/2$, both DWD and SVM are strongly inconsistent, i.e., their performances are the same as random guess; if $\gamma=1/2$, their performances are in-between. Therefore the signal size $\mu$ has to be large enough in order for SVM and DWD to gain some prediction power. This asymptotic study provides some useful information on the high dimensional behavior of the two classification methods but cannot be used to make quantitative comparison between them.

Recent rapid advances in statistical theory about the asymptotic performance of many classic machine learning algorithms in the limit of both large $n$ and $p$ have shed some light on this issue. There has been considerable effort to establish asymptotic results for different classification methods under the assumption that $p$ and $n$ grow at the same rate, that is, $n/p\rightarrow\alpha\textgreater 0$. The asymptotic results for SVM have been studied in \cite{Huang17} and \cite{maistatistical} under mixture models in which the data are assumed to be generated from a mixture distribution with two components, one for each class. The covariance matrix is assumed to follow a structure consisting of a pure background noise spiked with a few significant eigenvalues. The asymptotic results for DWD and logistic regression have been studied in \cite{8450750} and \cite{mailiao} respectively. In \cite{Huang17,8450750}, the spike eigenvector is assumed to be aligned with the signal direction while in \cite{maistatistical,mailiao} this assumption is relaxed. But all papers assume that the two classes have the same background noise. 

In the present work, we derive the asymptotic results for a general family of large-margin classifiers in the limit of $p,n\rightarrow\infty$ at fixed $\alpha=n/p$ under the two component mixture models. The family covers a broad range of margin-based classifiers including SVM, DWD, penalized logistic regression (PLR), and large-margin unified machine (LUM). The results in \cite{Huang17,8450750}, \cite{maistatistical}, \cite{mailiao,Mai2019HighDC}, and \cite{mailiao} are all special cases of this general result. We also consider more general settings in the sense that the signals are not necessarily aligned with the spiked eigenvectors and the background noises of two classes are not necessarily the same. We derive the analytical results using the replica method developed in statistical mechanics. All analytical results are confirmed by numerical experiments on finite-size systems and thus our formulas are verified to be correct. 
%To the best of our knowledge, the present paper is the first that provides not just bounds, but sharp predictions of the asymptotic behavior of the general margin-based classification estimators.  

%One important contribution of our analytical study is that it sheds light on how to select the best model and optimal tuning parameter for a given classification task. The selection method based on our theoretical analysis is proven to be much faster than traditional CV schemes. By comparing the asymptotic performances, we theoretically confirm that DWD outperforms SVM especially in HDLSS situations and therefore provide a solid analytic justification for the previous empirical phenomenon observed in \cite{Marron2007}. 

{\bf Related work}. The sharp asymptotics for hard margin SVM and unregularized logistic regression have been studied in \cite{montanari2019generalization} and \cite{candes2020phase} respectively under the single Gaussian models in which the data are assumed to be generated from a single Gaussian distribution. \cite{dobriban2018high} provide asymptotic analysis of the predictive risk of regularized discriminant analysis. \cite{dengmodel} studied hard margin SVM and unregularized logistic regression under Gaussian mixture models in which the data are assumed to be generated from Gaussian mixture distribution with two components, one for each class. \cite{gerace2020generalisation} studied the classification error for PLR and SVM for single Gaussian model with two layer neural network covariance structure. The analogous results for Gaussian mixture models with standard Gaussian components was provided in \cite{mignacco2020role}. \cite{wang2021binary} studied both max-margin SVM classifiers and min-norm interpolating classifiers under the popular generative Gaussian mixture model. The sharp asymptotics of generic convex generalized linear models was studied in \cite{gerbelot2020asymptotic} for rotationally invariant Gaussian data and in \cite{loureiro2021capturing} for block-correlated Gaussian data. The multi-class classification for mixture of Gaussians was also provided in \cite{loureiro2021learning} recently. Paralleling to classification, there has been considerable effort to establish the sharp asymptotics for regression. Examples include LASSO with i.i.d. setting \citep{bayati2011lasso}, LASSO with correlated data \citep{berthier2020state,celentano2020lasso}, ridgeless least squares \citep{hastie2019surprises}, generalized linear model \citep{Barbier5451}, and many others. 

Note that most of the results in aforementioned literature are rigorous under Gaussian assumption. The rigorous analysis methods include convex random geometry \citep{candes2020phase}, random matrix theorem \citep{dobriban2018high}, message-passing algorithms \citep{bayati2011lasso,berthier2020state,loureiro2021capturing,loureiro2021learning}, convex Gaussian min-max theorem \citep{montanari2019generalization,mignacco2020role,dengmodel}, and interpolation techniques \citep{Barbier5451}. The present work focuses on mixture of two component under spiked covariance setting without Gaussian assumption. While it remains an open problem to derive a rigorous proof for our results, we shall use simulation on moderate system sizes to provide numerical support that the theoretical formula is indeed exact in the high-dimensional limit.

The rest of this paper is organized as follows: In Section \ref{method}, we state the general framework for formulating the margin based classification methods. In Section \ref{perf}, the asymptotic results of the margin-based classifiers in the joint limit of large $p$ and $n$ for spiked population model are presented. Based on these asymptotic results, we study the separability phase transition in Section \ref{phase}. A method for estimating data parameters used in deriving the asymptotic results is provided in \ref{est}. In Section \ref{numeric}, we present numerical studies by comparing the theoretical results to Monte Carlo simulations on finite-size systems for several commonly used classification methods. An application of the proposed method to the breast cancer dataset is presented in Section \ref{real}. The last section is devoted to the conclusion.

\section{The Margin-Based Classification Method}\label{method}
In the binary classification problem, we are given a training dataset consisting of $n$ observations $\{(\bx_i,y_i); i=1,\cdots,n\}$ distributed according to some unknown joint probability distribution $P(\bx,y)$. Here $\bx_i\in\mR^p$ represents the input vector and $y_i\in\{+1,-1\}$ denotes the corresponding output class label, $n$ is the sample size, and $p$ is the dimension. There are $n_+$ and $n_-$ data in class $+$ and $-$ respectively. 

The goal of linear classification is to calculate a function $f(\bx)=\bx^T\bw+w_0$ such that sign$(f(\bx))$ can be used as the classification rule. Here $\bw\in\mR^p$ and $w_0\in\mR$ are parameters that need to be estimated. By definition of this classification rule, it is clear that correct classification occurs if and only if $yf(\bx)\textgreater 0$. Therefore, the quantity $yf(\bx)$, commonly referred as the functional margin, plays a critical role in classification techniques. The focus of this paper is on large-margin classification methods which can be fit in the regularization framework of Loss + Penalty. The loss function is used to keep the fidelity of the resulting model to the data while the penalty term in regularization helps to avoid overfitting of the resulting model. Using the functional margin, the regularization formulation of binary large-margin classifiers can be summarized as the following optimization problem
\begin{eqnarray}\label{class}
\min_{\bw,w_0}\left\{\sum_{i=1}^nV(y_i(\bx^T_i\bw+w_0))+\sum_{j=1}^pJ_\lambda(w_j)\right\},
\end{eqnarray}
where $V(\cdot)\ge 0$ is a loss function, $J_\lambda(\cdot)\ge 0$ is the regularization term, and $\lambda\textgreater 0$ is the tuning parameter for penalty. 

The general requirement for the loss function is convex decreasing with $V(u)\rightarrow 0$ as $u\rightarrow\infty$ and $V(u)\rightarrow\infty$ as $u\rightarrow -\infty$. Many commonly used classification techniques can be fit into this regularization framework. The examples include penalized logistic regression (PLR; \cite{lin2000}), support vector machine (SVM; \cite{Vapnik95}), distance weighted discrimination (DWD; \cite{Marron2007}), and large-margin unified machine (LUM; \cite{liu2011}). The loss functions of these classification methods are
\begin{eqnarray}\nn
\text{PLR}:&&V(u)=\log(1+\exp(-u)),\\\nn
\text{SVM}:&&V(u)=(1-u)_+,\\\nn
\text{DWD}:&&V(u)=\left\{\begin{array}{ccc}1-u&if&u\le \frac{q}{q+1}\\\frac{1}{u^q}\frac{q^q}{(q+1)^{q+1}}&if&u\textgreater\frac{q}{q+1}\end{array}\right.,\\\nn
\text{LUM}:&&V(u)=\left\{\begin{array}{ccc}1-u&if&u\le \frac{c}{1+c}\\\frac{1}{1+c}\left(\frac{a}{(1+c)u-c+a}\right)^a&if&u\textgreater\frac{c}{1+c}\end{array},\right.
\end{eqnarray}
where $q,a\textgreater 0$, and $c\ge 0$. It can be easily checked that SVM and DWD loss functions are special cases of the LUM loss function with appropriately chosen $a$ and $c$ \citep{liu2011}. For example, if we choose $a=c=q$, the LUM loss is the same as the DWD loss; if $a\textgreater 0$ and $c\rightarrow\infty$, the LUM loss is the same as the SVM loss. Besides the above methods, many other classification techniques can also be fit into the regularization framework, for example, the AdaBoost in Boosting \citep{FREUND1997119,friedman2000}, the import vector machine (IVM; \cite{doi:10.1198/106186005X25619}), and $\psi$-learning \citep{shen2003}. 

The commonly used penalty functions include $J_\lambda(w)=\frac{\lambda}{2}w^2$ for $L_2$ regularization and $J_\lambda(w)=\lambda|w|$ for sparse $L_1$ regularization. In this paper, we focus on the standard $L_2$ regularization.

\begin{figure}[hbtp]
	\begin{center}
		\epsfig{file=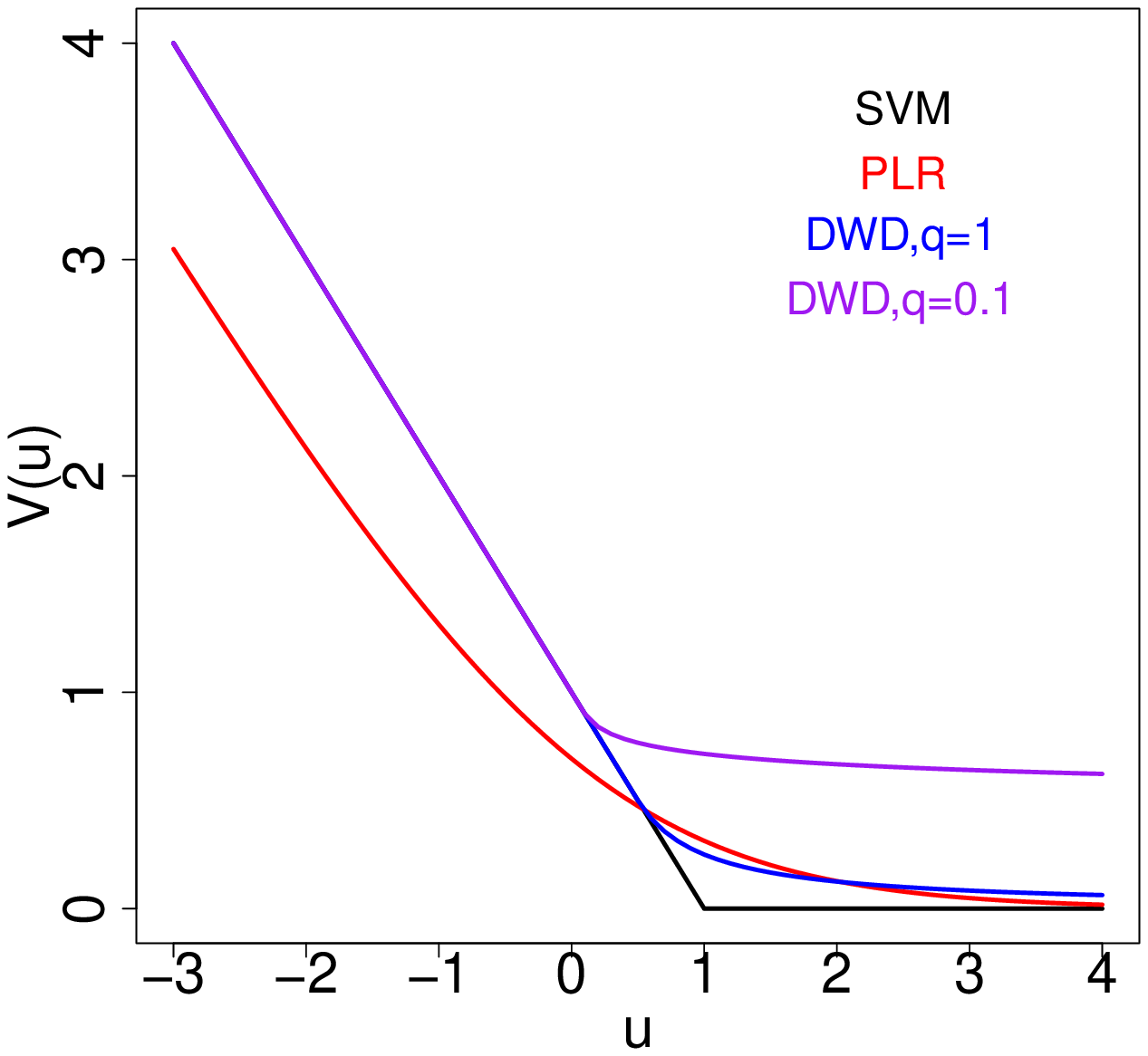,width=9.1cm,angle=-90}
	\end{center}
	\caption{Plots of various loss functions.}
	\label{figure1}
\end{figure}

Figure \ref{figure1} displays four loss functions: PLR, SVM, DWD with q=1, and DWD with q=0.1. Note that all loss functions have continuous first order derivatives except the hinge loss of SVM which is not differentiable at $u=1$. Among the four loss functions, PLR has all order derivatives while DWD only has first order derivative. As $u\rightarrow -\infty$, $V(u)\rightarrow -u$ for all methods.  As $u\rightarrow\infty$, $V(u)$ decays to 0 but with different speeds. The fastest one is SVM, followed by PLR, DWD with q=1, and DWD with q=0.1. We will see in Section \ref{numeric} that the decay speed of the loss function has a big influence on the classification performance in situations where $\lambda$ is small. Also all classification methods have the same performance when $\lambda$ is large enough due to the fact that $V(u)$ can be approximated by a linear function as $u\rightarrow 0$ for all loss functions.

\section{Asymptotic Performance}\label{perf}

Now let us specify the joint probability distribution $P(\bx,y)$. Conditional on $y=\pm 1$, assume that $\bx$ follows a multivariate distribution $P(\bx|y=\pm 1)$ with mean $\bmu_\pm$ and covariance matrix $\bSigma_\pm$. Here $\bmu_\pm\in{\mR}^p$ and $\bSigma_\pm$ denote a $p\times p$ positive definite matrix. Without loss of generality, we take $\bmu_+=\bmu$ and $\bmu_-=-\bmu$. 

We investigate the statistical behavior of the class separating hyperplane obtained from the optimization problem (\ref{class}) in the limit of $n,p\rightarrow\infty$ with $n/p\rightarrow\alpha$. Let us begin by introducing some notations. Denote $\bar{\bmu}=\bmu/\mu$, where $\mu=\|\bmu\|$. For a given loss function $V(u)$, define the proximal operator
\begin{eqnarray}\label{conj}
\psi(a,b)=\text{argmin}_u\left\{V(u)+\frac{(u-a)^2}{2b}\right\},
\end{eqnarray}
where $b\textgreater 0$. It can be considered as the solution of equation 
\begin{eqnarray}\nn
\partial V(u)+\frac{u-a}{b}=0,
\end{eqnarray}
where $\partial V(u)$ is one of the sub-gradients of $V(u)$. For convex $V(u)$, this equation has unique solution. Specifically, for SVM loss, we have closed form expression
\begin{eqnarray}\nn
\psi(a,b)=\left\{\begin{array}{ccc}a&if&a\ge 1\\1&if&1-b\le a\textless 1\\a+b&if&a\textless 1-b\end{array}\right..\\\nn
\end{eqnarray}
For DWD loss with $q=1$, we have
\begin{eqnarray}\nn
\psi(a,b)=\left\{\begin{array}{ccc}a+b&if&a\le 1/2-b\\\tilde{u}&if&a\textgreater 1/2-b\end{array}\right.,\\\nn
\end{eqnarray}
where $\tilde{u}$ is the solution of cubic equation $4u^3-4au^2-b=0$. For other loss functions, we have to rely on  certain numeric algorithms. Particularly for logistic loss, we can easily implement the Newton-Raphson algorithm because the loss function has closed form second order derivatives.

Our main results are based upon the following Claim for the distributional limit of estimators $\hat{\bw}$ obtained from (\ref{class}).
\begin{claim}\label{prop1}
The limiting distribution of $\hat{\bw}$ is the same as the limiting distribution of $\hat{\bmu}$ which is defined as
	\begin{eqnarray}\label{limitd}
	\hat{\bmu}=(\xi^+\bSigma_++\xi^-\bSigma_-+\lambda \bI_p)^{-1}\left(\sqrt{\xi_0^+}\bSigma_+^{1/2}\bz_++\sqrt{\xi_0^-}\bSigma_-^{1/2}\bz_-+\sqrt{p}\tilde{R}\bar{\bmu}\right), 
	\end{eqnarray}
	where $\bI_p$ is $p$-dimensional identity matrix and $\bz_\pm$ denote the vectors of length $p$ whose elements are i.i.d. standard Gaussian random variables independent of $\bmu$, and  
	\begin{eqnarray}\nn
	\xi^\pm=\frac{\alpha_\pm}{\sqrt{q^\pm_0}q^\pm}G_\pm,~
	\xi_0^\pm=\frac{\alpha_\pm}{(q^\pm)^2}H_\pm, \text{ and }
	\tilde{R}=\frac{\alpha_+\mu}{q^+}F_++\frac{\alpha_-\mu}{q^-}F_-.
	\end{eqnarray}
	Here $G_\pm$, $H_\pm$, and $F_\pm$ are functions of six quantities $q_0^\pm,q^\pm,R$, and $w_0$ defined as
	\begin{eqnarray}\nn
	F_\pm&=&E_z\left(\hat{u}_\pm-R\mu\mp w_0-\sqrt{q_0^\pm}z\right),\\\nn
	G_\pm&=&E_z\left\{\left(\hat{u}_\pm-R\mu\mp w_0-\sqrt{q_0^\pm}z\right)z\right\},\\\label{prime}
	H_\pm&=&E_z\left\{\left(\hat{u}_\pm-R\mu\mp w_0-\sqrt{q_0^\pm}z\right)^2\right\},
	\end{eqnarray}
	where $z$ is a standard Gaussian random variable and the expectation $E_z=\int\frac{dz}{\sqrt{2\pi}}\exp\left(-\frac{z^2}{2}\right)$. The $\hat{u}_\pm$ are also functions of $q_0^\pm,q^\pm,R$, and $w_0$ defined using (\ref{conj}) as 
	\begin{eqnarray}\nn
	\hat{u}_\pm&=&\psi\left(R\mu\pm w_0+\sqrt{q^\pm_0}z,q^\pm\right).
	\end{eqnarray}
	The values of $q_0^\pm,q^\pm,R,w_0$ can be obtained by solving the following six nonlinear equations:
	\begin{eqnarray}\label{eq1}
	q_0^\pm&=&\frac{1}{p}E_z(\hat{\bmu}^T\bSigma_\pm\hat{\bmu}),\\\label{eq2}
	R&=&\frac{1}{\sqrt{p}}E_z(\hat{\bmu}^T\bar{\bmu}),\\\label{eq3}
	\frac{\alpha_+}{q^+}F_+&=&\frac{\alpha_-}{q^-}F_-,\\\label{eq4}
	\frac{q^+}{\sigma_+^2}&=&\frac{q^-}{\sigma_-^2},\\\label{eq5}
	\frac{q^+\lambda}{\sigma_+^2}&=&1+\alpha_+G_++\alpha_-G_-.
	\end{eqnarray}
\end{claim}

From (\ref{limitd}), the limiting distribution of $\hat{\bw}$ is a multivariate normal with mean $(\xi^+\bSigma_++\xi^-\bSigma_-+\lambda \bI_p)^{-1}\left(\sqrt{p}\tilde{R}\bar{\bmu}\right)$ and covariance matrix $(\xi^+\bSigma_++\xi^-\bSigma_-+\lambda \bI_p)^{-1}\left(\xi_0^+\bSigma_++\xi_0^-\bSigma_-\right)(\xi^+\bSigma_++\xi^-\bSigma_-+\lambda \bI_p)^{-1}$, where all the parameters can be determined by solving a set of six nonlinear equations (\ref{eq1})-(\ref{eq5}). Note that two types of Gaussian random variables are introduced, i.e. primary variable $\hat{\bmu}$ and conjugate variable $\hat{u}$. The variances of these two random variables are controlled by $\xi_0^\pm$ and $q_0^\pm$ respectively. It is interesting to see that $\xi_0^\pm$ is determined by the expectation over a quadratic form of $\hat{u}$ i.e. (\ref{prime}), while $q_0^\pm$ is determined by the expectation over a quadratic form of $\hat{\bmu}$, i.e. (\ref{eq1}).  

The derivation of Claim \ref{prop1} is given in the Appendix based on the replica method developed in statistical mechanics. The replica method is a non-rigorous but highly sophisticated calculation procedure that has been used to derive a number of fascinating results in probability theory and information theory, see e.g. \cite{Tanaka,Guo,Verdu}. 
%So far, the rigorous verification of those results mainly focuses on `i.i.d. randomness', corresponding to the case $\bSigma_\pm=\bI_p$ here. Even in those simple cases, the proofs are rather challenging, see e.g. \cite{bayati}.

Using the asymptotic statistical behavior of the classification estimators provided in Claim \ref{prop1}, we are able to retrieve the asymptotic performance of the classification method (\ref{class}). Denote $\hat{\bw},\hat{w}_0$ the solution of (\ref{class}), the classification precision $P\{\pm(\bx_\pm^T\hat{\bw}+\hat{w}_0)\ge 0\}$ has an asymptotically deterministic behavior as given by the following Claim.
\begin{claim}\label{prop2}
	For $\bx_\pm$ generated from Class $\pm$, we have 
	\begin{eqnarray}\label{power}
	P\{\pm(\bx_\pm^T\hat{\bw}+\hat{w}_0)\ge 0\}\overset{P}{\rightarrow}\Phi(\zeta_\pm),
	\end{eqnarray}
	where $\Phi(\cdot)$ represents the cumulative distribution function of $N(0,1)$ and 
	\begin{eqnarray}\nn
	\zeta_\pm=\frac{R\mu\pm w_0}{\sqrt{q_0^\pm}}.
	\end{eqnarray}
	The values of the quantities $R$, $w_0$, and $q_0^\pm$ can be obtained from solving the equations  (\ref{eq1})-(\ref{eq5}) listed in Claim \ref{prop1}. 
\end{claim}
Claim \ref{prop2} allows us to assess the performance of different classification methods and obtain the value of $\lambda$ that yields the maximum precision for a given method. If we consider $\bSigma_+=\bSigma_-$ and PLR loss $V(u)$, Claims \ref{prop1} and \ref{prop2} end up with the results of \cite{mailiao}.

Now we consider datasets generated from the spiked covariance models which are particularly suitable for analyzing high dimensional statistical inference problems. Because for high dimensional data, typically only few components are scientifically important. The remaining structures can be considered as i.i.d. background noise. Therefore, we use a low-rank signal plus noise structure model \citep{ma2013,liu:sigclust}, and assume the following:
\begin{assump}\label{assum1}
	Each observation vector $\bx_+$ (resp. $\bx_-$) from Class $+1$ (resp. Class $-1$) can be viewed as an independent instantiation of the generative models
	\begin{eqnarray}\label{factor}
	\bx_+=\bmu+\sum_{k=1}^K\sigma_+\sqrt{\lambda^+_k}\bv_k\varepsilon_k+\bepsilon^+~~\left(resp.~~\bx_-=-\bmu+\sum_{k=1}^K\sigma_-\sqrt{\lambda^-_k}\bv_k\varepsilon_k+\bepsilon^-\right),
	\end{eqnarray} 
	where $\lambda_k^\pm\textgreater 0$ and $\bv_k\in\mR^p$ are orthonormal vectors for $k=1,\cdots,K$, i.e. $\bv_k^T\bv_k=1$ and $\bv_k^T\bv_{k^\prime}=0$ for $k\ne k^\prime$. The random variables $\varepsilon_1,\cdots,\varepsilon_K$ are i.i.d with mean 0 and variance 1. The elements of the p-vector $\bepsilon^\pm=\{\epsilon^\pm_1,\cdots,\epsilon^\pm_p\}$ are i.i.d random variables with $E(\epsilon^\pm_j)=0$, $E\{(\epsilon^\pm_j)^2\}=\sigma_\pm^2$, and $E\{(\epsilon^\pm_j)^3\}\textless\infty$. The $\epsilon^\pm_j$s and $\varepsilon_k$s are independent from each other. 
\end{assump}
In model (\ref{factor}), $\lambda^\pm_k$ represents the strength of the $k$-th signal component, and $\sigma_\pm^2$ represents the level of background noise. The real signal is typically low-dimensional, i.e. $K\ll p$. Here we use the most general assumption and allow different spiked covariances for different classes, e.g. we can have $\lambda^+_k=0$ and $\lambda^-_k\ne 0$. Note that the eigenvalue $\lambda^\pm_k$ is not necessarily decreasing in $k$ and $\lambda^\pm_1$ is not necessarily the largest eigenvalue. From (\ref{factor}), the covariance matrix becomes
\begin{eqnarray}\label{covariance}
\bSigma_\pm=\sigma_\pm^2\left(\bI_p+\sum_{k=1}^K\lambda^\pm_k\bv_k\bv_k^T\right).
\end{eqnarray} 
The $k$-th eigenvalue of $\bSigma_\pm$ is $\sigma_\pm^2(1+\lambda_k^\pm)$ for $k=1,\cdots,K$ and $\sigma_\pm^2$ for $k=K+1,\cdots,p$. Although the $\epsilon^\pm_j$s are i.i.d, we didn't impose any parametric form for the distribution of $\epsilon^\pm_j$ which allows for very flexible covariance structures for $\bx$, and thus the results are quite general. The requirement for the finite third order moment is to ensure Berry-Esseen central limit theorem applies. The Assumption \ref{assum1} is also called spiked population model and has been used in many situations, see \cite{Pastur,10.2307/2242400,citeulike:2714811,doi:10.1142/S0219024900000255,Johnstone,PhysRevLett.91.245701,Baik} for examples. 

Denote the projections of eigenvectors on the signal direction $\bar{\bmu}$ as $R_k=\bv_k^T\bar{\bmu}$ for $k=1,\cdots,K$; $R_{K+1}=\sqrt{1-\sum_{k=1}^KR_k^2}$; and $R_k=0$ for $k=K+2,\cdots,p$. After integration over $\bz_\pm$ in (\ref{eq1}) and (\ref{eq2}), we have the explicit formulas for $q_0^\pm$ and $R$ as 
\begin{eqnarray}\nn
q_0^\pm&=&\alpha_+H_++\alpha_-H_-+\left(\frac{\alpha_\pm\mu}{\sigma_\pm}F_\pm+\frac{\alpha_\mp\mu\sigma_\pm}{\sigma_\mp^2}F_\mp\right)^2\sum_{k=1}^{K+1}\frac{(1+\lambda_k^\pm)R_k^2}{\left\{1-\frac{\alpha_+\lambda^+_kG_+}{\sqrt{q_0^+}}-\frac{\alpha_-\lambda^-_kG_-}{\sqrt{q_0^-}}\right\}^2},\\\nn
R&=&\left(\frac{\alpha_+\mu}{\sigma_+}F_++\frac{\alpha_-\mu\sigma_+}{\sigma_-^2}F_-\right)\sum_{k=1}^{K+1}\frac{R_k^2}{\left\{1-\frac{\alpha_+\lambda^+_kG_+}{\sqrt{q_0^+}}-\frac{\alpha_-\lambda^-_kG_-}{\sqrt{q_0^-}}\right\}}.
\end{eqnarray}

Note that if $\sigma_+=\sigma_-$, we get $q^+=q^-$ directly from (\ref{eq4}). If we further assume $\bSigma_+=\bSigma_-$ and $\alpha_+=\alpha_-$, we get $w_0=0$, $q_0^+=q_0^-$, $q^+=q^-$ from (\ref{eq1}) and (\ref{eq3}). In this case, we only need to estimate three parameters and the results are much simpler.
\begin{claim}\label{prop3}
	 Under Assumption \ref{assum1}, assume that $\sigma_\pm$, $\lambda^\pm_k, R_k$, and $K$ are fixed irrespective of $p$. Further assume that $\bSigma_+=\bSigma_-=\bSigma$, $\alpha_+=\alpha_-=\alpha$, $\sigma_+=\sigma_-=\sigma$, and $\lambda_k^+=\lambda_k^-=\lambda_k$ for $k=1,\cdots,K$. Denote $q^+=q^-=q$ and $q_0^+=q_0^-=q_0$. Then the limiting distribution of $\hat{\bw}$ is the same as that of 
	\begin{eqnarray}\label{limitd0}
	(\xi\bSigma+\lambda \bI_p)^{-1}\left(\sqrt{\xi_0}\bSigma^{1/2}\bz+\sqrt{p}\tilde{R}\bar{\bmu}\right),
	\end{eqnarray}
	which leads to the asymptotic precision 
	\begin{eqnarray}\label{precision}
	P\{\pm(\bx_\pm^T\hat{\bw})\ge 0\}\rightarrow \Phi\left(\frac{R\mu}{\sqrt{q_0}}\right).
	\end{eqnarray}
	Here $\bz$ denotes the vectors of length $p$ whose elements are i.i.d. standard Gaussian random variables, and  
	\begin{eqnarray}\nn
	\xi=-\frac{\alpha}{\sqrt{q_0}q}f_2(q_0,q,R),~
	\xi_0=\frac{\alpha}{q^2}f_3(q_0,q,R), ~
	\tilde{R}=\frac{\alpha\mu}{q} f_1(q_0,q,R),
	\end{eqnarray}
	where the three functions $f_1,f_2,f_3$ are defined as
	\begin{eqnarray}\nn
	f_1(q_0,q,R)&=&E_z[(\hat{u}-R\mu-\sqrt{q_0}z)],\\\nn
	f_2(q_0,q,R)&=&E_z[(\hat{u}-R\mu-\sqrt{q_0}z)z],\\\nn
	f_3(q_0,q,R)&=&E_z[(\hat{u}-R\mu-\sqrt{q_0}z)^2].
	\end{eqnarray}
	Here the expectation is with respect to the standard Gaussian measure $z\sim N(0,1)$ and 
	\begin{eqnarray}\nn
	\hat{u}=\psi(R\mu+\sqrt{q_0}z,q).
	\end{eqnarray}
	The three parameters $q_0$, $q$, and $R$ are determined by the following three nonlinear equations
	\begin{eqnarray}\label{eq01}
	\frac{\sigma R}{\sqrt{q_0}}&=&\alpha\frac{\mu}{\sigma}\frac{f_1(q_0,q,R)}{\sqrt{q_0}}\sum_{k=1}^{K+1}\frac{R_k^2}{1-\alpha\lambda_kf_2(q_0,q,R)/\sqrt{q_0}},\\\label{eq02}
	1&=&\alpha\frac{f_3(q_0,q,R)}{q_0}+\left\{\alpha\frac{\mu}{\sigma}\frac{f_1(q_0,q,R)}{\sqrt{q_0}}\right\}^2\sum_{k=1}^{K+1}\frac{(1+\lambda_k)R_k^2}{\{1-\alpha\lambda_kf_2(q_0,q,R)/\sqrt{q_0}\}^2},\\\label{eq03}
	\frac{q\lambda}{\sigma^2}&=&1+\alpha\frac{f_2(q_0,q,R)}{\sqrt{q_0}}.
	\end{eqnarray}
\end{claim}
Note that for SVM loss $V(u)$, Claim \ref{prop3} is reduced to the results of \cite{Huang17} and \cite{maistatistical} while for DWD loss $V(u)$, it is reduced to the results of \cite{8450750}. Therefore, the results presented here are more general and can be used to compare the performance of different classification methods for a given problem. 

\section{Phase transition}\label{phase}

Based on the asymptotic results in Section \ref{perf}, in this section, we derive the phase transition for the non-regularized classification methods which solve the following optimization problem 
\begin{eqnarray}\label{nonreg}
\text{argmin}_{\bw\in{\mR}^p}\left\{\sum_{i=1}^nV(y_i\bx^T_i\bw)\right\}.
\end{eqnarray}
As shown in \cite{candes2020phase}, the existence for the non-regularized classification methods undergoes a phase transition, i.e. the solution of (\ref{nonreg}) does not exist in situations when the two classes of $n$ data points $(\bx_i,y_i)$ are completely linear separated and it does exist if the data points overlap. This is equivalent to establishing the the maximum number of training samples per dimensions below which the hard-margin SVM can have solution as shown in \cite{montanari2019generalization,9022461,dengmodel}. 

Consider the asymptotic regime where $n,p\rightarrow\infty$ such that 
\begin{eqnarray}\label{regime}
p/n\rightarrow\kappa\in(0,\tau], 
\end{eqnarray}
where $\tau\textgreater 0$ and $\kappa$ is called the overparametrization ratio. To quantify its effect on the test error, we study the problem of increasing dimensions as in (\ref{regime}) that further satisfy 
\begin{eqnarray}\label{signal}
\mu^2=s^2\kappa/\tau,&&\sigma^2=s^2(1-\kappa/\tau).
\end{eqnarray}
The following claim characterizes the phase transition of the model (\ref{nonreg}) in terms of $\kappa$ and $\tau$.
\begin{claim}\label{prop4}
Define $\kappa_{min}(\tau)$ as the solution of
\begin{eqnarray}\label{bound}
1&=&\frac{1}{\kappa}\int_{-\infty}^{z_c}(z_c-z)^2Dz+\frac{1}{\kappa(\tau-\kappa)}\sum_{k=1}^{K+1}\frac{R_k^2}{1+\lambda_k}\left\{\int_{-\infty}^{z_c}(z_c-z)Dz\right\}^2,
\end{eqnarray}
where $Dz=\frac{1}{\sqrt{2\pi}}\exp(-z^2/2)dz$ and $\Phi(z_c)=\kappa$, i.e. $z_c$ is the $\kappa$-th quantile of standard normal distribution. If the overparametrization ratio is smaller enough such that $\kappa\textless\kappa_{min}$, then the solution of equation (\ref{nonreg}) asymptotically exists with probability one. Conversely, if $\kappa\textgreater\kappa_{min}$, then the solution does not exist with probability one. 
\end{claim}
Note that Claim \ref{prop4} generalizes the result of \cite{dengmodel} for hard margin SVM which can be considered as a special case here if one chooses $\bSigma=\bI_p$, where $\bI_p$ is $p$-dimensional identity matrix.

\section{Estimation of data parameters}\label{est}
So far we assumed that the design covariance $\bSigma_\pm$ and other data parameters are known. In practice, we need to estimate $K,\mu, \sigma_\pm$, $\lambda^\pm_k$, and $R_k$ for $k=1,\cdots,K$ from the data. The problem of estimating covariance matrices in high-dimensional setting has attracted considerable attention in the past. Since the covariance estimation problem is not the focus of our paper, we will test the above approach using a simple covariance estimation method based the application of random matrix theory to spiked population model.

To estimate the background noise level $\sigma_\pm^2$, we use a robust variance estimate based on the full matrix of data values \citep{liu:sigclust}; that is, for the full set of $n_\pm p$ entries of the original $n_\pm\times p$ data matrix $\bX^\pm$, we calculate the robust estimate of scale, the median absolute deviation from the median (MAD), to estimate $\sigma_\pm$ as
\begin{eqnarray}\label{mad}
\hat{\sigma}_\pm&=&\frac{\text{MAD}_{\bX^\pm}}{\text{MAD}_{N(0,1)}}.
\end{eqnarray}
Here $\text{MAD}_{\bX^\pm}=\text{median}(|x^\pm_{ij}-\text{median}(\bX^\pm)|)$ and $\text{MAD}_{N(0,1)}=\text{median}(|r_{i}-\text{median}(\mR)|)$, where $\mR$ is a $n_\pm p$-dimensional vector whose elements are i.i.d. samples from $N(0,1)$ distribution. 

Denote $\bar{\bmu}_c=\bar{\bx}_+-\bar{\bx}_-$, where $\bar{\bx}_+=\frac{1}{n_+}\sum_{i=1}^{n_+}\bx_{+,i}$ and $\bar{\bx}_-=\frac{1}{n_-}\sum_{i=1}^{n_-}\bx_{-,i}$ represent the sample means for Class $+1$ and Class $-1$ respectively. Then, according to \cite{Huang17}, we estimate $\mu$ as
\begin{eqnarray}\nn
\hat{\mu}=\frac{1}{2}\sqrt{\|\bar{\bmu}_c\|^2-\frac{\hat{\sigma}_+^2}{\alpha_+}-\frac{\hat{\sigma}_-^2}{\alpha_-}}.
\end{eqnarray}

Denote $\tilde{\bSigma}_\pm$ the sample covariance matrix for Class $\pm 1$. Store all eigenvalues of $\tilde{\bSigma}_\pm$ greater than $(1+\sqrt{1/\alpha_\pm})^2-1$ as $[\tilde{\lambda}^\pm_1,\cdots,\tilde{\lambda}^\pm_{\hat{K}_\pm}]$ and their corresponding eigenvectors as $[\tilde{\bv}^\pm_1,\cdots,\tilde{\bv}^\pm_{\hat{K}_\pm}]$. Let $\hat{K}=\hat{K}_++\hat{K}_-$. By concatenating the spiked eigenvalues and eigenvectors from the two classes together, we obtain $\hat{K}$ spiked eigenvalues and their corresponding $\hat{K}$ eigenvectors. Then we relabel them and assign label $k\in[1,\cdots,\hat{K}_+]$ to Class +1 and label $k\in[\hat{K}_++1,\cdots,\hat{K}]$ to Class -1. To estimate $\lambda^\pm_k$ and $R_k$ for $k=1,\cdots,\hat{K}$, we use the results from \cite{Baik}. Define the function $P(u,v)=\sqrt{(1-1/uv^2)(1+1/uv)}$. For $k\in[1,\cdots,\hat{K}_+]$, we have
\begin{eqnarray}\nn
\hat{\lambda}^+_k&=&\frac{1}{2}\left(\tilde{\lambda}^+_k-\frac{1}{\alpha_+}+\sqrt{\left(\tilde{\lambda}^+_k-\frac{1}{\alpha_+}\right)^2-\frac{4}{\alpha_+}}\right)\\\label{baik1}
\tilde{R}_k&=&\frac{\bar{\bmu}_c^T\tilde{\bv}^+_k}{\|\bar{\bmu}_c\|P(\|\tilde{\bv}^+_k\|,\alpha_+)}
\end{eqnarray}
and $\hat{\lambda}^-_k=0$. For $k\in[\hat{K}_++1,\cdots,\hat{K}]$, we have
\begin{eqnarray}\nn
\hat{\lambda}^-_k&=&\frac{1}{2}\left(\tilde{\lambda}^-_{k-\hat{K}_+}-\frac{1}{\alpha_-}+\sqrt{\left(\tilde{\lambda}^-_{k-\hat{K}_+}-\frac{1}{\alpha_-}\right)^2-\frac{4}{\alpha_-}}\right)\\\label{baik2}
\tilde{R}_k&=&\frac{\bar{\bmu}_c^T\tilde{\bv}^-_{k-\hat{K}_+}}{\|\bar{\bmu}_c\|P(\|\tilde{\bv}^-_{k-\hat{K}_+}\|,\alpha_+)}
\end{eqnarray}
and $\hat{\lambda}^+_k=0$.

If we consider homogeneous situation where the two classes have the same covariance matrix, the results are much simpler. In this case, we need to combine two matrices $\bX_+$ and $\bX_-$ together to get a common set of spiked eigenvalues and eigenvectors. Then similar to (\ref{baik1}) and (\ref{baik2}), we use the results from \cite{Baik} to estimate the common eigenvalues $\lambda_k$ and projecting coefficients $R_k$.

\section{Numerical analysis}\label{numeric} 

In this section, we apply the general theoretical results derived in Section \ref{perf} to several specific classification methods by numerically solving the nonlinear equations for the corresponding loss functions. We aim to exploring and comparing different types of classifiers under various settings. Here we focus on homogeneous situations with $\bSigma_+=\bSigma_-$ and $\alpha_+=\alpha_-$ because in these situations the Bayes optimal classifiers are also linear and the classification performance can be exactly retrieved by the average precision derived in Claim \ref{prop2}. 
%Our main goal is to provide some guidelines on how to optimally choose classifiers and tuning parameters for a given dataset in practice. 

\begin{figure}[hbtp] \vspace{-0.3cm}
	\begin{center}
		\epsfig{file=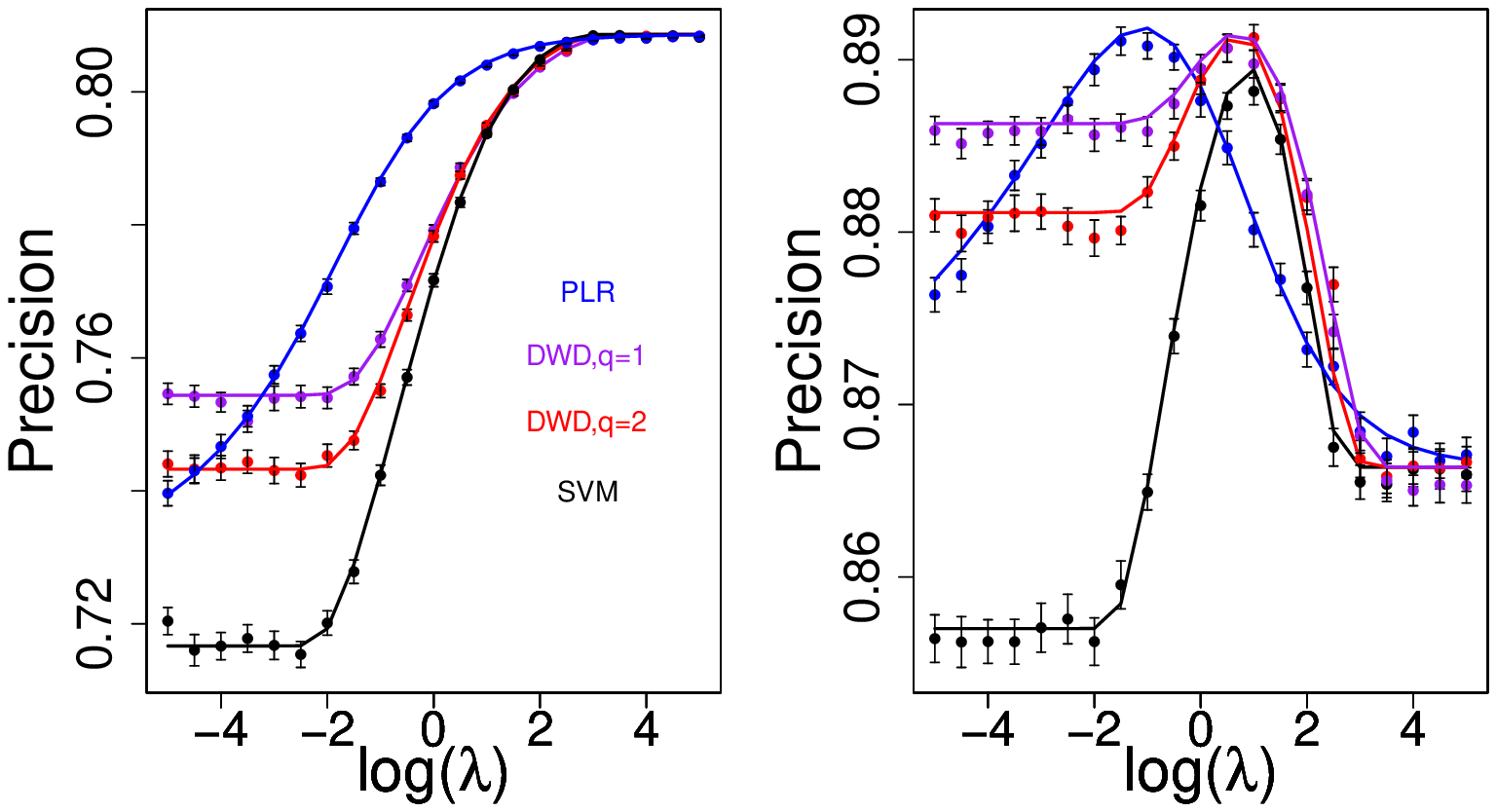,width=11.1cm,angle=-90}
	\end{center}
	\vspace{-1.5cm}
	\caption{Theoretical and empirical precision as a function of $\lambda$ for SVM,  DWD$_{q=1}$,  DWD$_{q=2}$, and PLR with $\alpha=1,\mu=2$. The empirical precision is taken by averaging over 100 simulated datasets with $p=500$. Left panel: the spike vector is aligned with $\mu$ with $K=2$, $\lambda_1=\lambda_2=4$, $R_1=1$, $R_2=0$. Right panel: the spike vector is not aligned with $\mu$ with $K=2$, $\lambda_1=\lambda_2=4$, $R_1=1/\sqrt{2}$, $R_2=0$}
	\label{figure2}
\end{figure}

To examine the validity of our analysis and to determine the finite-size effect, we first present some Monte Carlo simulations to confirm that our theoretical estimation derived in Section \ref{perf} is reliable. The performance of a classification method is assessed through the average precision computed based on (\ref{precision}).  Figures \ref{figure2} shows the comparison between our asymptotic estimations and simulations on finite dimensional datasets. We use the R packages $kernlab$, $glmnet$, $DWD$, and $DWDLargeR$ for solving SVM, PLR, DWD (q=1), and DWD (q=2) classification problem respectively. We didn't present simulation results for LUM and generalized DWD with non-integer $q$ because we cannot find reliable software package for solving this problem. The software package $DWDLargeR$ that is based on the algorithm developed in \cite{Defeng} does not provide the option for non-integer $q$. Here the dimension of the simulated data is $p = 500$ and the data are generated according to (\ref{factor}) in Assumption \ref{assum1} with i.i.d normal noise. We repeat the simulation 100 times for each parameter setting. The mean and standard errors over 100 replications are presented. 

From Figure \ref{figure2}, we can see that our analytical curves show fairly good agreement with the simulation experiment. Thus our analytical formula (\ref{precision}) provides reliable estimates for average precision even under moderate system sizes.

It is interesting to see that for small regularization parameter $\lambda$, the four patterns are quite different and SVM yields much smaller precision comparing to other three methods. On the other hand, if the tuning parameter $\lambda$ is large enough, the precision of all four methods approaches the same value. This is easy to understand because for large $\lambda$, the solution of (\ref{class}) is determined by the behavior of the loss function $V(u)$ at values of $u\rightarrow 0$ which turns out to be $1-u$ for SVM, DWD, LUM, and $\log(2)-u/2$ for PLR. Therefore, as $\lambda\rightarrow\infty$, the asymptotic results of  (\ref{class}) are approximately equal to the solution of 
\begin{eqnarray}\label{largelambda}
\text{argmin}_{\bw}\left\{\sum_{i=1}^n(c_1-c_2y_i\bx^T_i\bw)+\sum_{j=1}^p\frac{\lambda w_j^2}{2}\right\}
\end{eqnarray}
which is proportional to the weighted sample mean difference between two classes, i.e. \begin{eqnarray}\label{samplemean}
\hat{\bw}&\sim&\alpha_+\bar{\bx}_+-\alpha_-\bar{\bx}_-.
\end{eqnarray}
Here $c_1$ and $c_2$ are two constants, and $\bar{\bx}_+$ and $\bar{\bx}_-$ are the sample means for Class $+1$ and Class $-1$ respectively. On the other hand, for small $\lambda$, the solution of (\ref{class}) is also determined by the tail behavior of the loss function $V(u)$ at large $u$ values. Since the decay rates of different loss functions are quite different, this ends up with different behaviors at small $\lambda$ values as shown in Figures \ref{figure2}.

The difference between the settings of left panel and right panel of Figure \ref{figure2} is that in the left panel, the spike vectors $\bv_k$ $(k=1,\cdots,K)$ are either aligned with or orthogonal to $\bmu$ but in the right panel, the spike vectors are neither aligned nor orthogonal to $\bmu$. This discrepancy causes different patterns of the precision curves. In the left panel, the Bayes optimal solution is proportional to $\bmu$ which can be estimated using the difference of sample means between two classes. In this situation, as $\lambda$ increases, all solutions approach to the optimal one and thus we obtain increasing function for the precision. More specifically, it was pointed out in \cite{Huang17}, that the asymptotic value we can achieve for the precision is $\Phi\left(\frac{\rho_c}{\sqrt{1+\lambda_1\rho_c^2}}\frac{\mu}{\sigma}\right)$, where $\rho_c=\sqrt{\frac{\alpha\left(\frac{\mu}{\sigma}\right)^2}{1+\alpha\left(\frac{\mu}{\sigma}\right)^2}}$, and $\lambda_1$ represents the spiked eigenvalue in the $\bmu$ direction. In the right panel where $\bv_k$ is different in direction from $\bmu$, the Bayes optimal solution is proportional to $\bSigma^{-1}\bmu$, thus the asymptotic solution as $\lambda\rightarrow\infty$ is no long the optimal one. In this situation, we need to tune $\lambda$ so as to find the maximum precision for different methods. Note that this is consistent with the phenomenon observed in \cite{dobriban2018high,mignacco2020role} which show that in the case $\Sigma_\pm=I_p$ with balanced clusters, $\lambda=\infty$ gives the Bayes estimator, while in the unbalanced case the optimal regularization is finite. Because in balanced case, $\alpha_+=\alpha_-$ and (\ref{samplemean}) is the Bayes estimator of $\bmu$, while in unbalanced case, $\alpha_+\ne\alpha_-$ and (\ref{samplemean}) is not the Bayes estimator of $\bmu$.

\begin{figure}[hbtp] \vspace{-0.3cm}
		\begin{center}
			\epsfig{file=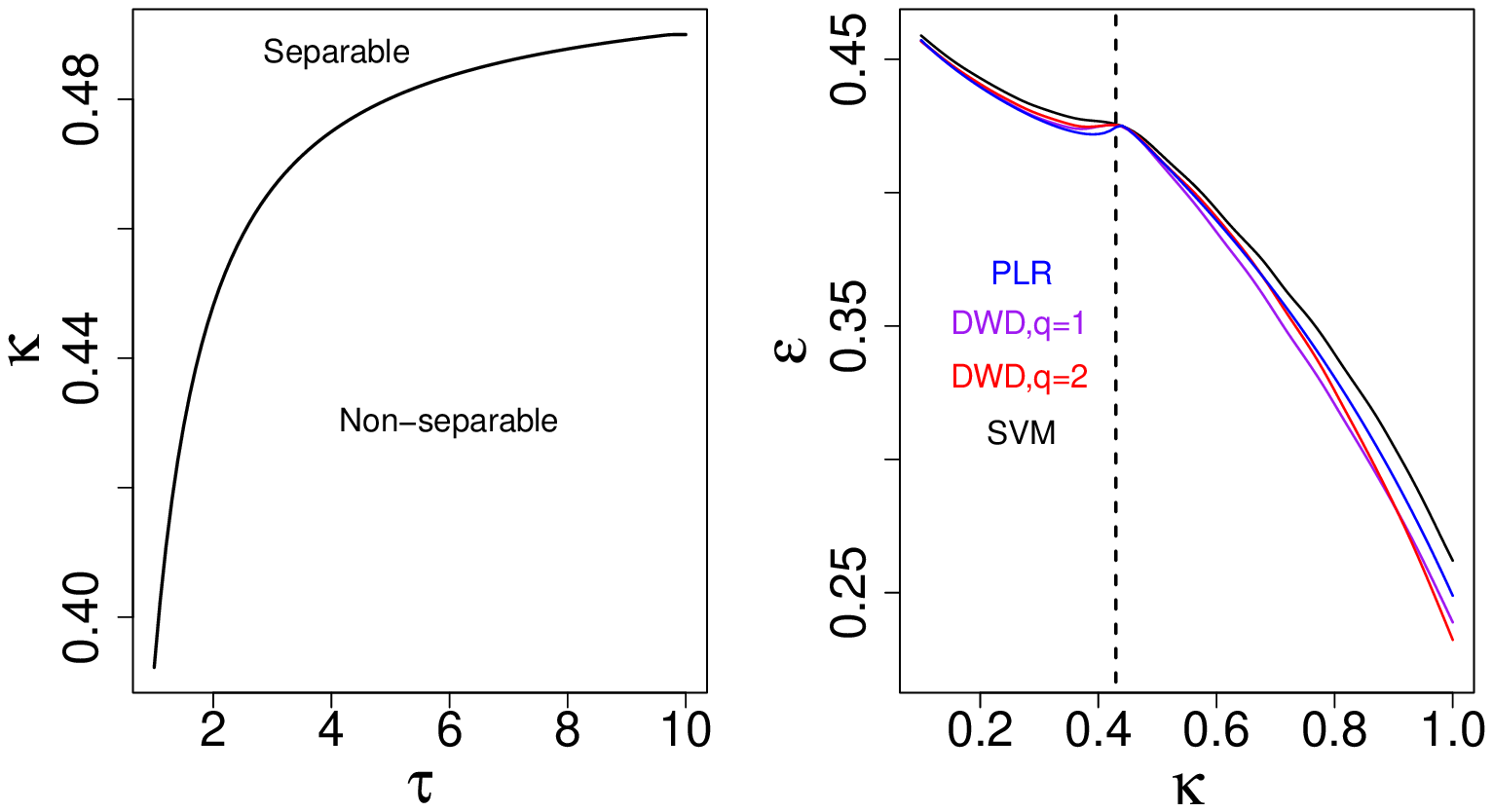,width=11.1cm,angle=-90}
		\end{center}
	\vspace{-1.5cm}
	\caption{Left panel: theoretical prediction for the phase transition curves. Right panel: theoretical classification error against the number of features per sample $\kappa=p/n$ for four classification methods with fixed $\tau=1.5$ under a small regularization setting $\lambda=10^{-5}$. Here $K=3$, $\lambda_1=\lambda_2=\lambda_3=4$, $R_1=R_2=1/2,R_3=0$. The vertical dashed line represents the  threshold $\kappa_{min}(1.5)$ of linear separability of the dataset.}
	\label{figure3}
\end{figure}

In Figure \ref{figure3}, we study the phase transition for the separability of two classes. The left panel of Figure \ref{figure3} displays the phase transition boundary for the separability of the two classes in the plane of $\kappa$ and $\tau$ which are defined in (\ref{regime}) and (\ref{signal}). Above the curve is the region where the probability of separating the two classes tends to 1 and below is the region where the probability of separating the two classes tends to 0. The prediction errors as a function of the overparametrization ratio $\kappa$ with fixed $\tau=1.5$ for the four classification methods under small regularization $\lambda=10^{-5}$ are shown in the right panel of Figure \ref{figure3}. The double descent behavior are found for all methods with peaks near the separability threshold $\kappa_{min}(1.5)$. This phenomenon indicates that the prediction error descends again after the threshold. A similar study has been given in \cite{dengmodel} for hard margin SVM and unregularized logistic regression under i.i.d covariance structure setting. The curves in the right panel of Figure \ref{figure3} also show that, as dimension increasing, DWD and PLR perform better than SVM under the non-regularized setting, i.e. $\lambda\rightarrow 0$. This is consistent with the previous empirical observations in \cite{Marron2007,Benito04}. The reason is that the small $\lambda$ behaviors of classification is determined by the decay speed of the corresponding loss function $V(u)$. The SVM hinge loss vanishes for the entire region of $u\ge 1$ but all the other loss functions decay to zero gradually as $u\rightarrow\infty$. 

\begin{figure}[hbtp] \vspace{-0.3cm}
		\begin{center}
			\epsfig{file=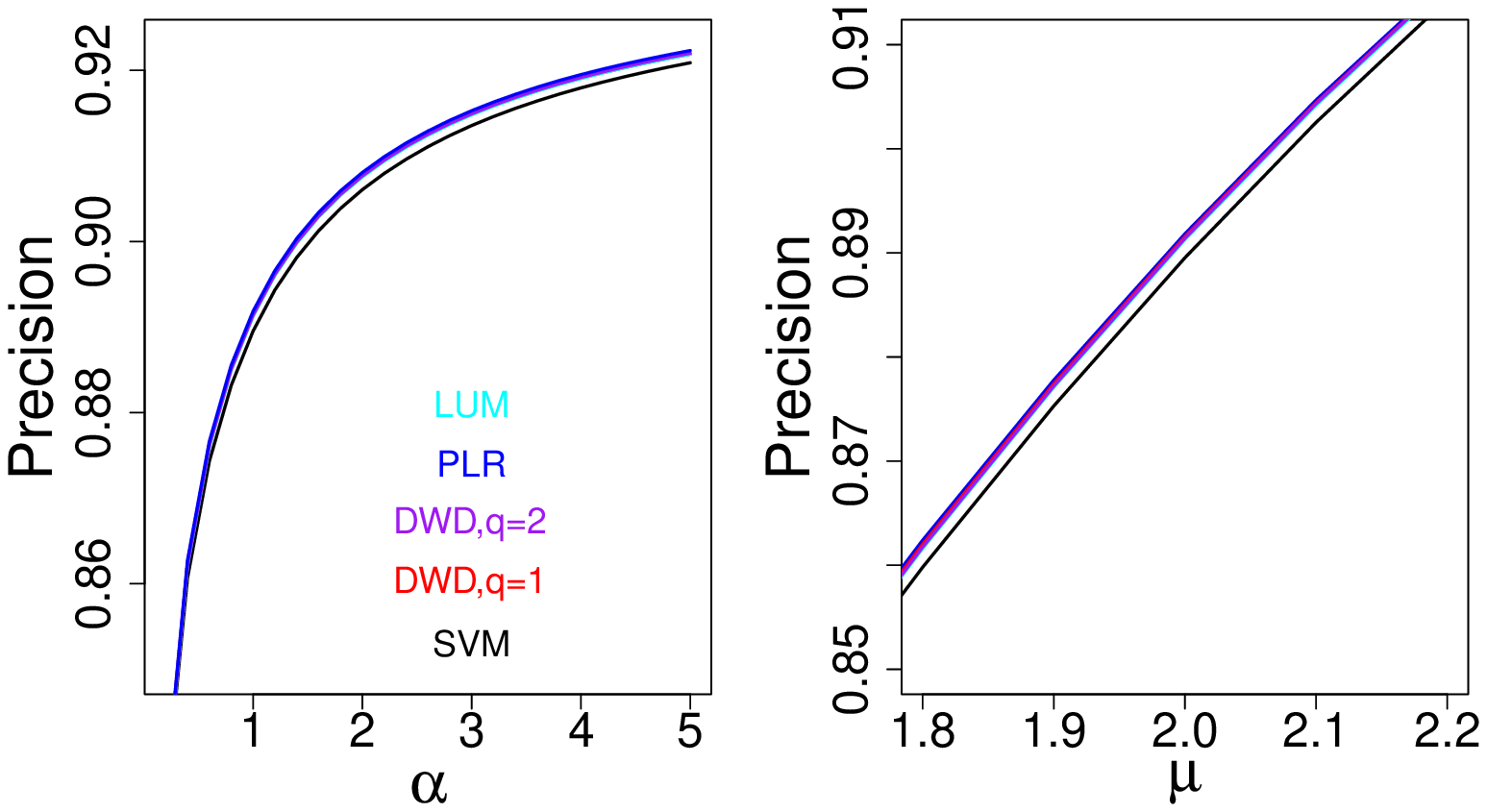,width=11.1cm,angle=-90}
		\end{center}
	\vspace{-1.5cm}
	\caption{Theoretical precision as a function of $\alpha$ and $\mu$ at optimal $\lambda$ for five different methods. Here $K=3$, $\lambda_1=\lambda_2=\lambda_3=4$, $R_1=R_2=1/2,R_3=0$. Left panel: theoretical precision as a function of $\alpha$ for fixed $\mu=2$. Right panel: theoretical precision as a function of $\mu$ for fixed $\alpha=1$.}
	\label{figure4}
\end{figure}

To further compare the performances of different methods, Figure \ref{figure4} plots the precision as a function of the parameters $\mu$ and $\alpha$ for optimal regularization, i.e. $\lambda$ is tuned to obtain the maximum precision. We consider five different methods which are SVM, PLR, DWD (q=1), DWD (q=2), and LUM (a=1,c=2). In Figure \ref{figure4}, the left panel plots the precision as a function of $\alpha$ with fixed $\mu$ and the right panel plots the precision as a function of $\mu$ with fixed $\alpha$. As it turns out, SVM performs worse than all the other four methods, but the discrepancy at optimal $\lambda$ is smaller than at small $\lambda$ as shown in Figure \ref{figure2}. The performance of all the other four methods are quite similar once $\lambda$ is optimally tuned.

We have performed numeric analysis under many other settings and the conclusions are quite similar. Overall, our analytical calculations agree well with the numerical simulations for moderate system sizes, and Claim \ref{prop2} provides reliable estimates for average precision. Our main observations from numeric analyses are
\begin{itemize}
	\item All methods achieve the same performances for large enough $\lambda$.
	\item For situations where the spiked vectors are the same in direction with $\bmu$, the optimal solutions of all methods are the same which are also equivalent to the limiting results as $\lambda\rightarrow\infty$.
	\item For situations where the spiked vectors are different in direction from $\bmu$, DWD and PLR are better than SVM especially when the regularization parameter $\lambda$ is small. This finding provides theoretical confirmations to the empirical results that have been observed in many previous simulation and real data studies.
	\item The previous empirical observation that DWD is better than SVM only holds at small $\lambda$. After carefully tuning for $\lambda$, the performance of all methods is quite similar. DWD, PLR, and LUM are slightly better than SVM at large $\alpha$.   
\item  The so-called double descent behaviors exist for all non-regularized margin-based classification methods with a peak at the separability threshold.
\end{itemize}
Note that the analytical demonstrations about the superior performance of  DWD over SVM at small $\lambda$ are consistent with many previous empirical findings. However, this does not mean that DWD is better than SVM at small $\lambda$ in all situations because our numerical results are derived based on the spiked population assumption which may not always hold in practice.

\section{Real Data}\label{real}

We apply our methods to a  breast cancer dataset from The Cancer Genome Atlas Research Network \citep{tcga2010} which include two subtypes: LumA and LumB. As in \cite{liu:sigclust,softthreshold}, we filter the genes using the ratio of the sample standard deviation and sample mean of each gene. After gene filtering, the dataset contained 235 patients with 169 genes. Among the 235 samples, there are 154 LumA samples and 81 LumB samples.

We consider LumA as Class +1 and LumB as Class -1. Assume the data are generated based on model (\ref{factor}), using the method discussed in Section \ref{est}, we  obtain the following parameter estimations: $\mu=4.81$, $\sigma=1.66$,  $\alpha=1.39$, $p=169$, $n=235$, $n_+=154$, $n_-=81$, $K=16$, $[\lambda_1,\cdots,\lambda_{16}]$ = [25.67, 12.00, 10.10,  9.37,  6.41,  4.92,  4.38,  4.19,  3.63,  3.09,  2.45,  1.96,1.87,  1.69,  1.57,  0.98], and $[R_1,\cdots, R_{16}]$ = [-0.67, -0.11,  0.57,  0.04,  0.18, -0.06, -0.03, -0.03, -0.16, -0.32,  0.01,  0.03,-0.06,  0.06,  0.03, -0.05]. 

\begin{figure}[hbtp]
	\begin{center}
		\epsfig{file=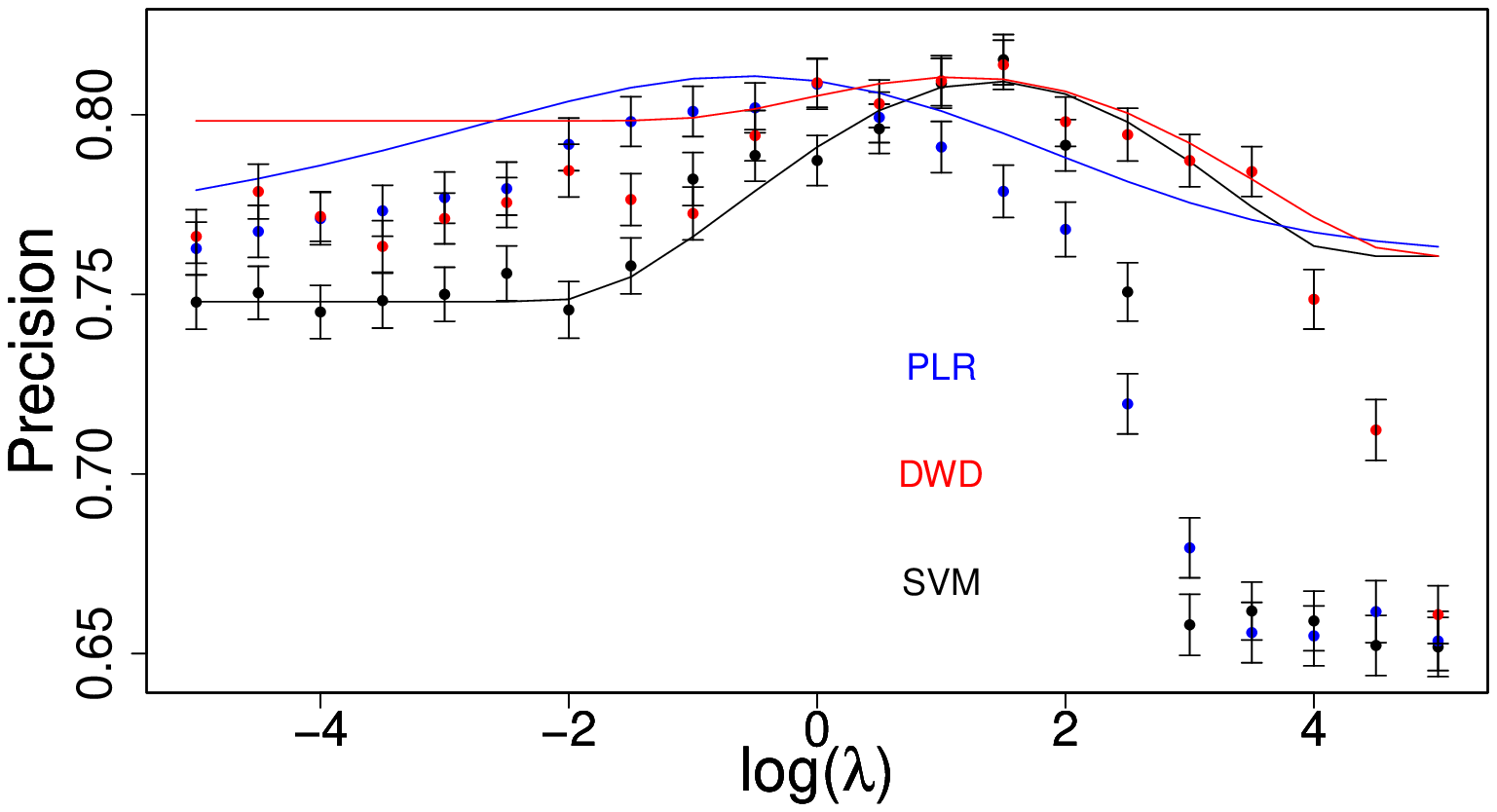,width=9.1cm,angle=-90}
	\end{center}
	\vspace{-1.5cm}
	\caption{The classification precision as a function of $\lambda$ based on theoretical estimation and cross validation analysis. The solid curves represent the theoretical results based on the parameters estimated from the breast cancer data. The error bars represent the cross validation experiment over 100 splittings.}
	\label{figure5}
\end{figure}

Figure \ref{figure5} plots the analytical curves of the average precision as functions of $\lambda$ for three classification methods SVM, PLR, and DWD(q=1). For comparison, the cross validation (CV) precision is computed by randomly splitting the data into two parts, 95\% for training and 5\% for test. The mean and standard deviation over 100 random splitting are presented. It can be seen that, at large $\lambda$, there are some discrepancies between the theoretical estimation and CV experiment. Note that, from (\ref{samplemean}), the solution of the margin based classification method (\ref{class}) can be approximated by the sample mean estimation if $\lambda$ is big enough. However, it is well known that the sample mean estimation method performs much worse than DWD and SVM if there are unbalanced sub-classes within each class as shown in \cite{liuxy}. Therefore, our results indicate that the data might include more complicated sub-cluster structure than the mixture of two simple components. On the other hand, at small $\lambda$, our theoretical estimation are quite close to the CV analysis. Particularly, at optimal $\lambda$, the theoretical estimation are 80.9\% (SVM), 81.0\% (DWD), 81.1\%(PLR) which are very close to the corresponding results based on CV analysis which are 81.5\% (SVM), 81.4\%(DWD), 80.9\%(PLR). Moreover, the maximum theoretical estimation for the three methods occurs at quite similar $\lambda$ values as the corresponding CV experiment. Overall, our theoretical results on the asymptotic precision can still provide reasonable guidelines on how to choose classification methods and tuning parameters for a given problem in practice.  

\section{Conclusion}\label{discussion}

Large-margin classifiers play an important role in classification problems. In this study, we examine the asymptotic behavior of a family of large-margin classifiers in the limit of $p,n\rightarrow\infty$ with fixed $\alpha=n/p$. This family includes many existing classifiers such as the SVM, DWD, PLR, and LUM as well as many new ones which can be built from the general convex loss function. Our focus is on the limiting distribution and classification precision of the estimators. On the basis of analytical evaluation, a method of selecting the best model and optimal tuning parameter is naturally developed for analyzing high dimensional data which significantly reduces the computational cost. Although our theoretical results are asymptotic in the problem dimensions, numerical simulations have shown that they are accurate already on problems with a few hundreds of variables.

Our analytical analyses provide deeper theoretical evidence to support the empirical conclusion that hard margin DWD yields better classification performance than hard margin SVM in high dimensions. Certainly, our observations may not be valid for all classification problems because we have applied the mixture of two components assumption with spiked covariance structure in numerical studies which cannot be true in all situations. Nevertheless, our analyses provide a convenient platform for deep investigation of the nature of margin-based classification methods and can also improve their practical use in various aspects as shown by the real data analysis in Section \ref{real}. Note that our numerical analysis focus on homogeneous cases where $\Sigma_+=\Sigma_-$. For non-homogeneous cases, the Bayes optimal solution is nonlinear and one possible solution is to use kernel based method. One of our future research topics is to derive the asymptotic behavior for the kernel based large margin classification methods. In situations where the spiked model cannot be applied or each class includes further sub-cluster structure, we plan to study the generalized spiked population model \citep{Bai2012167} or the classification methods that can incorporate the sub-cluster analysis. 

\section*{Acknowledgements} 
The authors thank the editor and three referees for many helpful comments and suggestions which led to a much improved presentation. This research is supported in part by Division of Mathematical Sciences (National Science Foundation) Grant DMS-1916411 (Huang).

\appendix
\section*{Appendix}
\label{sec1}
\setcounter{equation}{0}
\def\theequation{A\arabic{equation}}

This appendix outlines the replica calculation leading to Claims 1. Claims 2 and 3 are just direct applications of Claim 1. We limit ourselves to the main steps. For a general introduction to the replica method and its motivation, we refer to \cite{mezard1987spin,mezard2009information}. 

Denote $\bX=[\bx_1,\cdots,\bx_n]^T$, $\by=(y_1,\cdots,y_n)^T$. Among the $n$ samples, let the first $n_+$ ones belong to Class +1, i.e. $y_i=1$ for $i\in\{1,\cdots,n_+\}$ and the last $n_-$ ones belong to Class -1, i.e. $y_i=-1$ for $i\in\{n_++1,\cdots,n\}$. We consider regularized classification of the form
\begin{eqnarray}\label{classreg}
(\hat{\bw},\hat{w}_0)&=&\text{argmin}_{\bw,w_0}\left\{\sum_{i=1}^nV\left(\frac{y_i\bx_i^T\bw}{\sqrt{p}}+y_iw_0\right)+\sum_{j=1}^pJ_\lambda(w_j)\right\}.
\end{eqnarray}
After suitable scaling, the terms inside the bracket $\{\cdot\}$ are exactly equal to the objective function of model (\ref{class}) in the main text. 

The replica calculation aims at estimating the following moment generating function ( partition function)
\begin{eqnarray}\nn
&&Z_{\beta}(\bX,\by)\\\nn
&=&\int\exp\left\{-\beta\left[\sum_{i=1}^nV\left(\frac{y_i\bx_i^T\bw}{\sqrt{p}}+y_iw_0\right)+\sum_{j=1}^pJ_\lambda(w_j)\right]\right\}d\bw dw_0\\\nn
&=&\int\exp\left\{-\beta\left[\sum_{i=1}^{n_+}V\left(\frac{\bx_i^T\bw}{\sqrt{p}}+w_0\right)+\sum_{i=n_++1}^{n}V\left(-\frac{\bx_i^T\bw}{\sqrt{p}}-w_0\right)\right.\right.\\\label{zbeta}
&&\left.\left.~~~~~~~~~~~~~~~~~~~~~~~~~~~~~~~~~~~~~~~~+\sum_{j=1}^pJ_\lambda(w_j)\right]\right\}d\bw dw_0,
\end{eqnarray}
where $\beta\textgreater 0$ is a `temperature' parameter. In the zero temperature limit, i.e. $\beta\rightarrow\infty$, $Z_{\beta}(\bX,\by)$ is dominated by the values of $\bw$ and $w_0$ which are the solution of (\ref{classreg}). 

Within the replica method, it is assumed that the limits $p\rightarrow\infty$, $\beta\rightarrow\infty$ exist almost surely for the quantity $(p\beta)^{-1}\log Z_{\beta}(\bX,\by)$, and that the order of the limits can be exchanged. We therefore define the free energy 
\begin{eqnarray}\label{calf1}
{\cal F}&=&-\lim_{\beta\rightarrow\infty}\lim_{p\rightarrow\infty}\frac{1}{p\beta}\log Z_{\beta}(\bX,\by)=-\lim_{p\rightarrow\infty}\lim_{\beta\rightarrow\infty}\frac{1}{p\beta}\log Z_{\beta}(\bX,\by).
\end{eqnarray}
Notice that, by (\ref{calf1}) and using Laplace method in the integral (\ref{zbeta}), we have
\begin{eqnarray}\nn
{\cal F}&=&\lim_{p\rightarrow\infty}\frac{1}{p}\min_{\bw,w_0}\left\{\sum_{i=1}^nV\left(\frac{y_i\bx_i^T\bw}{\sqrt{p}}+y_iw_0\right)+\sum_{j=1}^pJ_\lambda(w_j)\right\}.
\end{eqnarray}
It is also assumed that $p^{-1}\log Z_{\beta}(\bX,\by)$ concentrates tightly around its expectation so that the free energy can in fact be evaluated by computing
\begin{eqnarray}\label{calf}
{\cal F}&=&-\lim_{\beta\rightarrow\infty}\lim_{p\rightarrow\infty}\frac{1}{p\beta}\left\langle\log Z_{\beta}(\bX,\by)\right\rangle_{\bX,\by},
\end{eqnarray}
where the angle bracket stands for the expectation with respect to the distribution of training data $\bX$ and $\by$. 

In order to evaluate the integration of a log function, we make use of the replica method based on the identity 
\begin{eqnarray}
\log Z=\lim_{k\rightarrow 0}\frac{\partial Z^k}{\partial k}=\lim_{k\rightarrow 0}\frac{\partial}{\partial k}\log Z^k,
\end{eqnarray}
and rewrite (\ref{calf}) as
\begin{eqnarray}\label{replica}
{\cal F}=-\lim_{\beta\rightarrow\infty}\lim_{p\rightarrow\infty}\frac{1}{p\beta}\lim_{k\rightarrow 0}\frac{\partial}{\partial k}\Xi_k(\beta),
\end{eqnarray}
where 
\begin{eqnarray}\label{xi0}
\Xi_k(\beta)=\langle\{Z_\beta(\bX,\by)\}^k\rangle_{\bX,\by}=\int\{Z_\beta(\bX,\by)\}^k\prod_{i=1}^nP(\bx_i,y_i)d\bx_i dy_i.
\end{eqnarray}
Equation (\ref{replica}) can be derived by using the fact that $\lim_{k\rightarrow 0}\Xi_k(\beta)=1$ and exchanging the order of 
the averaging and the differentiation with respect to $k$. In the replica method, we will first evaluate $\Xi_k(\beta)$ for integer $k$ and then 
apply to real $k$ and take the limit of $k\rightarrow 0$. 

For integer $k$, in order to represent $\{Z_\beta(\bX,\by)\}^k$ in the integrand of (\ref{xi0}), we use the identity
\begin{eqnarray}\nn
\left(\int f(x)\nu(dx)\right)^k=\int f(x_1)\cdots f(x_k)\nu(dx_1)\cdots\nu(dx_k),
\end{eqnarray}
where $\nu(dx)$ denotes the measure over $x\in\mR$. We obtain
\begin{eqnarray}\nn
\{Z_\beta(\bX,\by)\}^k&=&\prod_{a=1}^k\left[\int \exp\left\{-\beta\left[\sum_{i=1}^nV\left(\frac{y_i\bx_i^T\bw^a}{\sqrt{p}}+y_iw_0^a\right)\right.\right.\right.\\\label{zd}
&&~~~~~~~~~~~~~~~~~~~~~~~~~~~\left.\left.\left.+\sum_{j=1}^pJ_\lambda(w_j^a)\right]\right\}d\bw^a dw_0^a\right]
\end{eqnarray}
where we have introduced replicated parameters
\begin{eqnarray}\nn
\bw^a\equiv[w^a_1,\cdots,w^a_p]^T&\text{and}&w_0^a, \text{ for }a=1,\cdots,k.
\end{eqnarray}
Exchanging the order of the two limits $p\rightarrow\infty$ and $k\rightarrow 0$ in (\ref{replica}), we have 
\begin{eqnarray}\label{freef}
{\cal F}=-\lim_{\beta\rightarrow\infty}\frac{1}{\beta}\lim_{k\rightarrow 0}\frac{\partial}{\partial k}\left(\lim_{p\rightarrow\infty}\frac{1}{p}\Xi_k(\beta)\right).
\end{eqnarray}
Define the measure $\nu(d\bw)$ over $\bw\in\mathbb{R}^p$ as follows
\begin{eqnarray}\nn
\nu(d\bw)&=&\int\exp\left\{-\beta\sum_{j=1}^pJ_\lambda(w_j)\right\}d\bw.
\end{eqnarray}
Similarly, define the measure $\nu_+(d\bx)$ and $\nu_-(d\bx)$ over $\bx\in\mathbb{R}^p$ as
\begin{eqnarray}\nn
\nu_+(d\bx)=P(\bx|y=+1)d\bx&\text{ and }&\nu_-(d\bx)=P(\bx|y=-1)d\bx. 
\end{eqnarray}
In order to carry out the calculation of $\Xi_k(\beta)$, we let $\nu^k(d\bw)\equiv\nu(d\bw^1)\times\cdots\times\nu(d\bw^k)$ be a measure over $(\mathbb{R}^p)^k$, with $\bw^1,\cdots,\bw^k\in\mathbb{R}^p$. Analogously $\nu^n(d\bx)\equiv\nu(d\bx_1)\times\cdots\times\nu(d\bx_n)$ with $\bx_1,\cdots,\bx_n\in\mathbb{R}^p$, $\nu^n(dy)\equiv\nu(dy_1)\times\cdots\times\nu(dy_n)$ with $y_1,\cdots,y_n\in\{-1,1\}$, and $\nu^k(dw_0)\equiv\nu(dw_0^1)\times\cdots\times\nu(dw_0^k)$ with $w_0^1,\cdots,w_0^k\in\mathbb{R}$. With these notations, we have
\begin{eqnarray}\nn
\Xi_k(\beta)&=&\int\exp\left\{-\beta\sum_{i=1}^n\sum_{a=1}^kV\left(\frac{y_i\bx_i^T\bw^a}{\sqrt{p}}+y_iw_0^a\right)\right\}\nu^k(d\bw)\nu^k(dw_0)\nu^n(d\by)\nu^n(d\bx)\\\nn
&=&\int\left[\int\exp\left\{-\beta\sum_{a=1}^kV\left(\frac{\bx^T\bw^a}{\sqrt{p}}+w_0^a\right)\right\}\nu_+(d\bx)\right]^{n_+}\times\\\nn
&&~~~\left[\int\exp\left\{-\beta\sum_{a=1}^kV\left(\frac{-\bx^T\bw^a}{\sqrt{p}}-w_0^a\right)\right\}\nu_-(d\bx)\right]^{n_-}\nu^k(d\bw)\nu^k(dw_0)\\\label{xi}
&=&\int\exp\{p(\alpha_+\log I_++\alpha_-\log I_-)\}\nu^k(d\bw)\nu^k(dw_0),
\end{eqnarray}
where $\alpha_\pm=n_\pm/p$ and
\begin{eqnarray}\label{iterm}
I_\pm&=&\int\exp\left\{-\beta\sum_{a=1}^kV\left(\frac{\pm\bx^T\bw^a}{\sqrt{p}}\pm w_0^a\right)\right\}\nu_\pm(d\bx).
\end{eqnarray}
Notice that above we used the fact that the integral over $(\bx_1,\cdots,\bx_n)\in(\mathbb{R}^p)^n$ factors into $n_+$ integrals over $(\mathbb{R})^p$ with measure $\nu_+(d\bx)$ and $n_-$ integrals over $(\mathbb{R})^p$ with measure $\nu_-(d\bx)$. We next use the identity 
\begin{eqnarray}\label{identity}
e^{f(x)}&=&\frac{1}{2\pi}\int^{\infty}_{-\infty}\int^{\infty}_{-\infty}e^{i\left(q-x\right)\tilde{Q}+f(q)}dqd\tilde{Q}.
\end{eqnarray}
We apply this identity to (\ref{iterm}) and introduce integration variables $du^a,d\tilde{u}^a$ for $1\le a\le k$. Letting $\nu^k(du)=du^1\cdots du^k$ and $\nu^k(d\tilde{u})=d\tilde{u}^1\cdots d\tilde{u}^k$
\begin{eqnarray}\nn
I_\pm&=&\int\exp\left\{-\beta\sum_{a=1}^kV\left(u^a\pm w_0^a\right)+i\sqrt{p}\sum_{a=1}^k\left(u^a\mp\frac{\bx^T\bw^a}{\sqrt{p}}\right)\tilde{u}^a\right\}\nu_\pm(d\bx)\nu^k(du)\nu^k(d\tilde{u})\\\nn
&=&\int\exp\left\{-\beta\sum_{a=1}^kV\left(u^a\pm w_0^a\right)+i\sqrt{p}\sum_{a=1}^ku^a\tilde{u}^a-\frac{1}{2}\sum_{ab}(\bw^a)^T\bSigma_\pm\bw^b\tilde{u}^a\tilde{u}^b\right.\\\label{iplus}
&&~~~~~~~~~~~~~~~~~~~~~~~~~~~\left.-i\sum_{a=1}^k(\bw^a)^T\bmu\tilde{u}^a\right\}\nu^k(du)\nu^k(d\tilde{u}).
\end{eqnarray}
Note that, conditional on $y=\pm 1$, $\bx$ follows multivariate distributions with mean $\pm\bmu$ and covariance matrices $\bSigma_\pm$. In deriving (\ref{iplus}), we have used the fact that the low-dimensional marginals of $\bx$ can be approximated by Gaussian distribution based on multivariate central limit theorem.  

Next we apply (\ref{iplus}) to (\ref{xi}), and introduce integration variables $Q^\pm_{ab},\tilde{Q}^\pm_{ab}$ and $R^a,\tilde{R}^a$ associated with $(\bw^a)^T\bSigma_\pm\bw^b/p$ and $(\bw^a)^T\bar{\bmu}/\sqrt{p}$ respectively for $1\le a,b\le k$. Denote $\bw_0=(w_0^a)_{1\le a\le k}$, $\bQ_\pm\equiv(Q_{ab}^\pm)_{1\le a,b\le k}$, $\tilde{\bQ}_\pm\equiv(\tilde{Q}_{ab}^\pm)_{1\le a,b\le k}$, $\vR\equiv(R^a)_{1\le a\le k}$, and $\tilde{\vR}\equiv(\tilde{\vR}^a)_{1\le a\le k}$. Note that, constant factors can be applied to the integration variables, and we choose convenient factors for later calculations.  Letting $d\bQ^\pm\equiv\prod_{a,b}dQ_{ab}^\pm$ $d\tilde{\bQ}_\pm\equiv\prod_{a,b}d\tilde{Q}_{ab}^\pm$, $d\vR\equiv\prod_{a}dR^{a}$, and $d\tilde{\vR}\equiv\prod_{a}d\tilde{R}^{a}$, we obtain
\begin{eqnarray}\label{saddle}
\Xi_k(\beta)=\int\exp\left\{-p{\cal S}_k(\bQ_\pm,\tilde{\bQ}_\pm,\vR,\tilde{\vR},\bw_0)\right\}d\bQ_{+}d\bQ_{-}d\tilde{\bQ}_{+}d\tilde{\bQ}_{-}d\vR d\tilde{\vR}\nu^k(dw_0),
\end{eqnarray}
where
\begin{eqnarray}\nn
{\cal S}_k(\bQ_\pm,\tilde{\bQ}_\pm,\vR,\tilde{\vR},\bw_0)&=&-i\beta\left(\sum_{ab}Q_{ab}^+\tilde{Q}_{ab}^++\sum_{ab}Q_{ab}^-\tilde{Q}_{ab}^-+\sum_aR^{a}\tilde{R}^{a}\right)\\\nn
&&-\frac{1}{p}\log\xi(\tilde{\bQ}_\pm,\tilde{\vR})-\hat{\xi}(\bQ_\pm,\vR,\bw_0),\\\nn
\xi(\tilde{\bQ}_\pm,\tilde{\vR})&=&\int\exp\left\{-i\beta\sum_{ab}\tilde{Q}_{ab}^+(\bw^a)^T\bSigma_+\bw^b-i\beta\sum_{ab}\tilde{Q}_{ab}^-(\bw^a)^T\bSigma_-\bw^b\right.\\\nn
&&~~~~~~~~~~~~~~~~~~\left.-i\beta\sum_a\sqrt{p}\tilde{R}^a(\bw^a)^T\bar{\bmu}\right\}\nu^k(d\bw),\\\label{xihat}
\hat{\xi}(\bQ_\pm,\vR,\bw_0)&=&\alpha_+\log\hat{I}_++\alpha_-\log\hat{I}_-,
\end{eqnarray}
where
\begin{eqnarray}\nn
\hat{I}_\pm&=&\int\exp\left\{-\beta\sum_{a=1}^kV\left(u^a\pm w_0^a\right)+i\sqrt{p}\sum_{a=1}^ku^a\tilde{u}^a\right.\\\label{ihat}
&&~~~~~~~~~~\left.-\frac{p}{2}\sum_{ab}Q^\pm_{ab}\tilde{u}^a\tilde{u}^b-i\sqrt{p}\sum_{a=1}^kR^a\mu\tilde{u}^a\right\}\nu^k(du)\nu^k(d\tilde{u}).
\end{eqnarray}
Now we apply steepest descent method to the remaining integration. According to Varadhan's Claim \citep{Tanaka}, only the saddle points of the exponent of the integrand contribute to the integration in the limit of $p\rightarrow\infty$. We next use the saddle point method in (\ref{saddle}) to obtain
\begin{eqnarray}\nn
-\lim_{p\rightarrow\infty}\frac{1}{p}\Xi_k(\beta)&=&{\cal S}_k(\bQ_\pm^\star,\tilde{\bQ}_\pm^\star,\vR^\star,\tilde{\vR}^\star,\bw_0^\star),
\end{eqnarray}
where $\bQ_\pm^\star,\tilde{\bQ}_\pm^\star,\vR^\star,\tilde{\vR}^\star,\bw_0^\star$ is the saddle point location. Looking for saddle-points over all the entire space is in general difficult to perform. We  assume replica symmetry for saddle-points such that they are invariant under exchange of any two replica indices $a$ and $b$, where $a\ne b$. Under this symmetry assumption, the space is greatly reduced and the exponent of the integrand can be explicitly evaluated.  The replica symmetry is also motivated by the fact that ${\cal S}_k(\bQ_\pm^\star,\tilde{\bQ}_\pm^\star,\vR^\star,\tilde{\vR}^\star,\bw_0^\star)$ is indeed left unchanged by such change of variables. This is equivalent to postulating that $(w_0^a)^\star=w_0$, $R^a=R$, $\tilde{R}^a=i\tilde{R}$, 
\begin{eqnarray}\label{qsym}
(Q_{ab}^\pm)^\star=\left\{\begin{array}{cc}q^\pm_1&\text{if a=b}\\q^\pm_0&\text{otherwise}\end{array}\right.,&\text{and}&(\tilde{Q}_{ab}^\pm)^\star=\left\{\begin{array}{cc}i\frac{\beta\zeta_1^\pm}{2}&\text{if a=b}\\i\frac{\beta\zeta_0^\pm}{2}&\text{otherwise}\end{array}\right.,
\end{eqnarray}
where the factor $i\beta/2$ is for future convenience. The next step consists in substituting the above expressions for $\bQ_\pm^\star,\tilde{\bQ}_\pm^\star,\vR^\star,\tilde{\vR}^\star,\bw_0^\star$ in ${\cal S}_k(\bQ_\pm^\star,\tilde{\bQ}_\pm^\star,\vR^\star,\tilde{\vR}^\star,\bw_0^\star)$ and then taking the limit $k\rightarrow 0$. We will consider separately each term of ${\cal S}_k(\bQ_\pm^\star,\tilde{\bQ}_\pm^\star,\mR^\star,\tilde{R}^\star,\bw_0^\star)$. Let us begin with the first term
\begin{eqnarray}\nn
&&-i\beta\left(\sum_{ab}Q_{ab}^+\tilde{Q}_{ab}^++\sum_{ab}Q_{ab}^-\tilde{Q}_{ab}^-+\sum_aR^a\tilde{R}^{a}\right)\\\label{qqhat}
&=&\frac{k\beta^2}{2}(\zeta^+_1q^+_1-\zeta^+_0q_0^+)+\frac{k\beta^2}{2}(\zeta^-_1q^-_1-\zeta^-_0q_0^-)+k\beta R\tilde{R}.
\end{eqnarray}
Next consider $\log\xi(\tilde{\bQ}_\pm,\tilde{\vR})$. For p-vectors $\bu,\bv\in\mR^p$ and $p\times p$ matrix $\bSigma$, introducing the notation $\|\bv\|_{\bSigma}^2\equiv \bv^T\bSigma\bv$ and $\langle\bu,\bv\rangle\equiv\sum_{j=1}^pu_jv_j/p$, we have
\begin{eqnarray}\nn
\xi(\tilde{\bQ}_\pm,\tilde{\vR})&=&\int\exp\left\{\frac{\beta^2}{2}(\zeta^+_1-\zeta^+_0)\sum_{a=1}^k\|\bw^a\|^2_{\bSigma_+}+\frac{\beta^2\zeta^+_0}{2}\sum_{a,b=1}^k(\bw^a)^T\bSigma_+\bw^b\right.\\\nn
&&~~~~~~~~~+\frac{\beta^2}{2}(\zeta^-_1-\zeta^-_0)\sum_{a=1}^k\|\bw^a\|^2_{\bSigma_-}+\frac{\beta^2\zeta^-_0}{2}\sum_{a,b=1}^k(\bw^a)^T\bSigma_-\bw^b\\\nn
&&~~~~~~~~~\left.+\beta\sqrt{p}\sum_{a=1}^k\tilde{R}(\bw^a)^T\bar{\bmu}\right\}\nu^k(d\bw)\\\nn
&=&E\int\exp\left\{\frac{\beta^2}{2}(\zeta^+_1-\zeta^+_0)\sum_{a=1}^k\|\bw^a\|^2_{\bSigma_+}+\beta\sqrt{\zeta^+_0}\sum_{a=1}^k(\bw^a)^T\bSigma_+^{1/2}\bz_+\right.\\\nn
&&~~~~~~~~~+\frac{\beta^2}{2}(\zeta^-_1-\zeta^-_0)\sum_{a=1}^k\|\bw^a\|^2_{\bSigma_-}+\beta\sqrt{\zeta^-_0}\sum_{a=1}^k(\bw^a)^T\bSigma_-^{1/2}\bz_-\\\label{xiq}
&&~~~~~~~~~\left.+\beta\sqrt{p}\sum_{a=1}^k\tilde{R}(\bw^a)^T\bar{\bmu}\right\}\nu^k(d\bw),
\end{eqnarray}
where expectation is with respect to $\bz_+,\bz_-\sim N(0,I_{p\times p})$. Notice that, given $\bz_+,\bz_-\in \mR^p$, the integrals over $\bw^1,\cdots,\bw^k$ factorize, whence
\begin{eqnarray}\nn
\xi(\tilde{\bQ}_\pm,\tilde{\vR})&=&E\left\{\left[\int\exp\left\{\frac{\beta^2}{2}(\zeta^+_1-\zeta^+_0)\|\bw\|^2_{\bSigma_+}+\beta\sqrt{\zeta^+_0}\bw^T\bSigma_+^{1/2}\bz_+\right.\right.\right.\\\nn
&&+\frac{\beta^2}{2}(\zeta^-_1-\zeta^-_0)\|\bw\|^2_{\bSigma_-}+\beta\sqrt{\zeta^-_0}\bw^T\bSigma_-^{1/2}\bz_-\\\nn
&&\left.\left.\left.+\beta\sqrt{p}\tilde{R}\bw^T\bar{\bmu}\right\}\nu^k(d\bw)\right]^k\right\}.
\end{eqnarray}
Finally, after integration over $\nu^k(d\tilde{u})$, (\ref{ihat}) becomes
\begin{eqnarray}\nn
\hat{I}_\pm&=&\int\exp\left\{-\beta\sum_{a=1}^kV\left(u^a\pm w_0\right)-\frac{1}{2}\sum_{ab}(u^a-R\mu)(\bQ_\pm^{-1})_{ab}(u^b-R\mu)\right.\\\label{ihat1}
&&~~~~~~~~~\left.-\frac{1}{2}\log\text{det}\bQ_\pm\right\}\nu^k(du).
\end{eqnarray}

We can next take the limit $\beta\rightarrow\infty$. The analysis of the saddle point parameters $q^\pm_0,q^\pm_1,\zeta^\pm_0,\zeta^\pm_1$ shows that $q^\pm_0,q^\pm_1$ have the same limit with $q^\pm_1-q^\pm_0=(q^\pm/\beta)+o(\beta^{-1})$ and $\zeta^\pm_0,\zeta^\pm_1$ have the same limit with $\zeta^\pm_1-\zeta^\pm_0=(-\zeta^\pm/\beta)+o(\beta^{-1})$. Substituting the above expression in (\ref{qqhat}) and (\ref{xiq}), in the limit of $k\rightarrow 0$, we then obtain
\begin{eqnarray}\nn
&&-i\beta\left(\sum_{ab}Q_{ab}^+\tilde{Q}_{ab}^++\sum_{ab}Q_{ab}^-\tilde{Q}_{ab}^-+\sum_aR^a\tilde{R}^{a}\right)\\\label{first}
&=&\frac{k\beta}{2}(\zeta^+_0q^+-\zeta^+q_0^+)+\frac{k\beta}{2}(\zeta^-_0q^--\zeta^-q_0^-)+k\beta R\tilde{R},
\end{eqnarray}
and
\begin{eqnarray}\nn
\xi(\tilde{\bQ}_\pm,\tilde{\vR})&=&E\left\{\left[\int\exp\left\{-\frac{\beta\zeta^+}{2}\|\bw\|^2_{\bSigma_+}+\beta\sqrt{\zeta^+_0}\bw^T\bSigma_+^{1/2}\bz_+\right.\right.\right.\\\nn
&&~~~~~~~~~~~~~~~-\frac{\beta\zeta^-}{2}\|\bw\|^2_{\bSigma_-}+\beta\sqrt{\zeta^-_0}\bw^T\bSigma_-^{1/2}\bz_-\\\label{second}
&&~~~~~~~~~~~~~~~\left.\left.\left.+\beta\sqrt{p}\tilde{R}\bw^T\bar{\bmu}\right\}\nu^k(d\bw)\right]^k\right\}.
\end{eqnarray}
Similarly, using (\ref{qsym}), we obtain
\begin{eqnarray}\nn
\sum_{ab}(u^a-R\mu)(\bQ_\pm^{-1})_{ab}(u^b-R\mu)&=&\frac{\beta\sum_a(u^a-R\mu)^2}{q^\pm}-\frac{\beta^2q^\pm_0\{\sum_a(u^a-R\mu)\}^2}{(q^\pm)^2},\\\nn
\log\text{det}\bQ_\pm&=&\log\left[(q^\pm_1-q^\pm_0)^k\left(1+\frac{kq^\pm_0}{q^\pm_1-q^\pm_0}\right)\right]=\frac{k\beta q^\pm_0}{q^\pm},
\end{eqnarray}
where we retain only the leading order terms. Therefore, (\ref{ihat1}) becomes
\begin{eqnarray}\nn
\hat{I}_\pm&=&\exp\left(-\frac{k\beta q^\pm_0}{2q^\pm}\right)\int Dz_\pm\left(\int\exp\left\{-\beta V(u\pm w_0)-\frac{\beta(u-R\mu-\sqrt{q^\pm_{0}}z_\pm)^2}{2q^\pm}\right.\right.\\\nn
&&~~~~~~~~~~~~~~~~~~~~~~~~~~~~~~~~~~~~~~~~\left.\left.+\frac{\beta q^\pm_{0}z_\pm^2}{2q^\pm}\right\}du\right)^k,
\end{eqnarray}
where the expectation $D_z=\int\frac{dz}{\sqrt{2\pi}}\exp\left(-\frac{z^2}{2}\right)$. Substituting this expression in (\ref{xihat}), we obtain
\begin{eqnarray}\nn
\hat{\xi}(\bQ_\pm,\mR,\bw_0)&=&-k\beta E\left\{\alpha_+\min_u\left[V(u+w_0)+\frac{(u-R\mu-\sqrt{q^+_{0}}z_+)^2}{2q^+}\right]\right.\\\label{third}
&&~~~~~~~~~\left.+\alpha_-\min_u\left[V(u-w_0)+\frac{(u-R\mu-\sqrt{q^-_{0}}z_-)^2}{2q^-}\right]\right\},
\end{eqnarray}
where the expectation is with respect to $z_+,z_-\sim N(0,1)$. Putting (\ref{first}), (\ref{second}), and (\ref{third}) together into (\ref{saddle}) and then into (\ref{replica}), we obtain 
\begin{eqnarray}\nn
{\cal F}&=&\frac{1}{2}(\zeta^+_0q^+-\zeta^+q^+_0)+\frac{1}{2}(\zeta^-_0q^--\zeta^-q^-_0)+R\tilde{R}\\\nn
&&+\alpha_+\text{E}\min_{u\in R}\left\{V(u+w_0)+\frac{\left(u-R\mu-\sqrt{q^+_0} z_+\right)^2}{2q^+}\right\}\\\nn
&&+\alpha_-\text{E}\min_{u\in R}\left\{V(u-w_0)+\frac{\left(u-R\mu-\sqrt{q^-_0} z_-\right)^2}{2q^-}\right\}\\\nn
&&+\frac{1}{p}\text{E}\min_{\bw\in R^p}\left\{\frac{\zeta^+}{2}\|\bw\|_{\bSigma_+}^2+\frac{\zeta^-}{2}\|\bw\|_{\bSigma_-}^2
-\left\langle\sqrt{\zeta_0^+}\bSigma_+^{1/2}\bz_++\sqrt{\zeta_0^-}\bSigma_-^{1/2}\bz_-+\sqrt{p}\tilde{R}\bar{\bmu},\bw\right\rangle\right.\\\label{fs}
&&\left.~~~~~~~~~~~~~+\sum_{j=1}^pJ_\lambda(w_j)\right\},
\end{eqnarray}
where the expectations are with respect to $ z_+,z_-\sim N(0,1)$, and $\bz_+,\bz_-\sim N(0,\bI_{p\times p})$, with $z_+,z_-$ and $\bz_+,\bz_-$ independent from each other. Here $\zeta^\pm,\zeta_0^\pm,q^\pm,q_0^\pm,R,\tilde{R}$ are order parameters which can be determined from the saddle point equations of ${\cal F}$. Define the functions $F_\pm$, $G_\pm$, and $H_\pm$ as 
\begin{eqnarray}\nn
F_\pm&=&E_z\left(\hat{u}_\pm-R\mu\mp w_0-\sqrt{q_0^\pm}z_\pm\right),\\\nn
G_\pm&=&E_z\left\{\left(\hat{u}_\pm-R\mu\mp w_0-\sqrt{q_0^\pm}z_\pm\right)z\right\},\\\nn
H_\pm&=&E_z\left\{\left(\hat{u}_\pm-R\mu\mp w_0-\sqrt{q_0^\pm}z_\pm\right)^2\right\},
\end{eqnarray}
where  
\begin{eqnarray}\nn
\hat{u}_\pm&=&\text{argmin}_{u\in R}\left\{V(u\pm w_0)+\frac{(u-R\mu\mp w_0-\sqrt{q^\pm_0}z_\pm)^2}{2q^\pm}\right\}.
\end{eqnarray}
The result in (\ref{fs}) is for general penalty function $J_\lambda(w)$. For quadratic penalty $J_\lambda(w)=\lambda w^2$, we get the closed form limiting distribution of $\bw$ as 
\begin{eqnarray}\label{wlimit}
\hat{\bw}&=&(\xi^+\bSigma_++\xi^-\bSigma_-+\lambda \bI_p)^{-1}\left(\sqrt{\xi_0^+}\bSigma_+^{1/2}\bz_++\sqrt{\xi_0^-}\bSigma_-^{1/2}\bz_-+\sqrt{p}\tilde{R}\bar{\bmu}\right).
\end{eqnarray}
All the order parameters can be determined by the following saddle-point equations:
\begin{eqnarray}\label{eq001}
\xi_0^\pm&=&\frac{\alpha_\pm}{(q^\pm)^2}H_\pm,\\\label{eq002}
\xi^\pm&=&\frac{\alpha_\pm G_\pm}{\sqrt{q^\pm_0}q^\pm},\\\label{eq003}
q_0^\pm&=&\frac{1}{p}E_z\|\bw\|^2_{\bSigma_\pm},\\\label{eq04}
q^\pm&=&\frac{1}{p\sqrt{\zeta_0^\pm}}E\left\langle\bSigma_\pm^{1/2}\bz_\pm,\hat{\bw}\right\rangle\\\label{eq05}
R&=&\frac{1}{\sqrt{p}}E_z\langle\bar{\bmu},\hat{\bw}\rangle,\\\label{eq06}
\tilde{R}&=&\frac{\alpha_+\mu}{q^+}F_++\frac{\alpha_-\mu}{q^-}F_-,\\\label{eq07}
\frac{\alpha_+}{q^+}F_+&=&\frac{\alpha_-}{q^-}F_-.
\end{eqnarray}
The above formulas are for general positive definite covariance matrix $\bSigma_\pm$. Then after applying the spiked population assumption (2) and integrating over $\bz_\pm$, we obtain the explicit nonlinear equations for determining six parameters $q_0^\pm,q^\pm,R,w_0$ as
\begin{eqnarray}\nn
q_0^\pm&=&\alpha_+H_++\alpha_-H_-+\left(\frac{\alpha_\pm\mu}{\sigma_\pm}F_\pm+\frac{\alpha_\mp\mu\sigma_\pm}{\sigma_\mp^2}F_\mp\right)^2\sum_{k=1}^{K+1}\frac{(1+\lambda_k^\pm)R_k^2}{\left\{1-\frac{\alpha_+\lambda^+_kG_+}{\sqrt{q_0^+}}-\frac{\alpha_-\lambda^-_kG_-}{\sqrt{q_0^-}}\right\}^2},\\\nn
R&=&\left(\frac{\alpha_+\mu}{\sigma_+}F_++\frac{\alpha_-\mu\sigma_+}{\sigma_-^2}F_-\right)\sum_{k=1}^{K+1}\frac{R_k^2}{\left\{1-\frac{\alpha_+\lambda^+_kG_+}{\sqrt{q_0^+}}-\frac{\alpha_-\lambda^-_kG_-}{\sqrt{q_0^-}}\right\}}
,\\\nn
\frac{\alpha_+}{q^+}F_+&=&\frac{\alpha_-}{q^-}F_-,\\\nn
\frac{q^+}{\sigma_+^2}&=&\frac{q^-}{\sigma_-^2},\\\nn
\frac{q^+\lambda}{\sigma_+^2}&=&1+\alpha_+G_++\alpha_-G_-.
\end{eqnarray}
Then, the other five parameters $\zeta_0^\pm$, $\zeta^\pm$, and $\tilde{R}$ can be obtained using equations (\ref{eq001}), (\ref{eq002}), and (\ref{eq06}).  

\section*{Derivation of Claim \ref{prop4}}

Let $\lambda=0$, from equations (\ref{eq01}), (\ref{eq02}), and (\ref{eq03}) in Claim \ref{prop3}, we obtain
\begin{eqnarray}\nn
q_0-\frac{R^2}{\gamma^2}&=&\alpha E\{(\hat{u}-a)^2\},\\\nn
\frac{R}{\gamma^2}&=&\alpha\mu E(\hat{u}-a),\\\nn
1&=&-\frac{\alpha}{\sqrt{q_0}}E\{(\hat{u}-a)z\},
\end{eqnarray}
where $a=R\mu+\sqrt{q_0}z$, $\hat{u}=\psi(a,q)$, and $\gamma^2=\hat{\bmu}^T\bSigma^{-1}\hat{\bmu}$. For SVM, define $z_c=(1-R\mu)/\sqrt{q_0}$, $x=q/\sqrt{q_0}$, and $r=R/\sqrt{q_0}$, we have
\begin{eqnarray}\label{m1}
1-\frac{r^2}{\gamma^2}&=&\alpha\left\{\int_{z_c-x}^{z_c}(z_c-z)^2Dz+x^2\int_{-\infty}^{z_c-x}Dz\right\}\\\label{m2}
\frac{r}{\gamma^2}&=&\alpha\mu\left\{\int_{z_c-x}^{z_c}(z_c-z)Dz+x\int_{-\infty}^{z_c-x}Dz\right\}\\\label{m3}
1&=&\alpha\int_{z_c-x}^{z_c}Dz.
\end{eqnarray}
(\ref{m1}) and (\ref{m2}) lead to
\begin{eqnarray}\nn
1&=&\alpha\left\{\int_{z_c-x}^{z_c}(z_c-z)^2Dz+x^2\int_{-\infty}^{z_c-x}Dz\right\}+\left\{\alpha\gamma\mu\left(\int_{z_c-x}^{z_c}(z_c-z)Dz+x\int_{-\infty}^{z_c-x}Dz)\right)\right\}^2.
\end{eqnarray}
For fixed $\alpha$, $\mu$ has upper bound in order for the above equation to have a solution. Because of (\ref{m3}), the biggest value for $\mu$ we can achieve is when $x\rightarrow\infty$. Therefore the phase transition is determined by
\begin{eqnarray}\nn
1&=&\alpha\int_{-\infty}^{z_c}(z_c-z)^2Dz+\left\{\alpha\gamma\mu\int_{-\infty}^{z_c}(z_c-z)Dz\right\}^2,
\end{eqnarray}
where $\Phi(z_c)=1/\alpha$. Note that $\psi=1/\alpha$, substituting the spike covariance matrix (\ref{covariance}) and (\ref{signal}), we obtain (\ref{bound}).
$\hfill\blacksquare$

\bibliographystyle{chicago} 
\bibliography{biblist}

\begin{thebibliography}{}

\bibitem[\protect\citeauthoryear{Bai and Yao}{Bai and Yao}{2012}]{Bai2012167}
Bai, Z. and J.~Yao (2012).
\newblock On sample eigenvalues in a generalized spiked population model.
\newblock {\em Journal of Multivariate Analysis\/}~{\em 106}, 167 -- 177.

\bibitem[\protect\citeauthoryear{Baik and Silverstein}{Baik and
  Silverstein}{2006}]{Baik}
Baik, J. and J.~W. Silverstein (2006).
\newblock Eigenvalues of large sample covariance matrices of spiked population
  models.
\newblock {\em Journal of Multivariate Analysis\/}~{\em 97\/}(6), 1382 -- 1408.

\bibitem[\protect\citeauthoryear{Barbier, Krzakala, Macris, Miolane, and
  Zdeborov{\'a}}{Barbier et~al.}{2019}]{Barbier5451}
Barbier, J., F.~Krzakala, N.~Macris, L.~Miolane, and L.~Zdeborov{\'a} (2019).
\newblock Optimal errors and phase transitions in high-dimensional generalized
  linear models.
\newblock {\em Proceedings of the National Academy of Sciences\/}~{\em
  116\/}(12), 5451--5460.

\bibitem[\protect\citeauthoryear{Bayati and Montanari}{Bayati and
  Montanari}{2011}]{bayati2011lasso}
Bayati, M. and A.~Montanari (2011).
\newblock The lasso risk for gaussian matrices.
\newblock {\em IEEE Transactions on Information Theory\/}~{\em 58\/}(4),
  1997–2017.

\bibitem[\protect\citeauthoryear{Benito, Parker, Du, Skoog, Lindblom, Perou,
  and Marron}{Benito et~al.}{2004}]{Benito04}
Benito, M., J.~Parker, Q.~Du, L.~Skoog, A.~Lindblom, C.~M. Perou, and J.~S.
  Marron (2004).
\newblock Adjustment of systematic microarray data biases.
\newblock {\em Bioinformatics\/}~{\em 20}, 105--144.

\bibitem[\protect\citeauthoryear{Berthier, Montanari, and Nguyen}{Berthier
  et~al.}{2020}]{berthier2020state}
Berthier, R., A.~Montanari, and P.-M. Nguyen (2020).
\newblock State evolution for approximate message passing with non-separable
  functions.
\newblock {\em Information and Inference: A Journal of the IMA\/}~{\em 9\/}(1),
  33–79.

\bibitem[\protect\citeauthoryear{Candès, Sur, et~al.}{Candès
  et~al.}{2020}]{candes2020phase}
Candès, E.~J., P.~Sur, et~al. (2020).
\newblock The phase transition for the existence of the maximum likelihood
  estimate in high-dimensional logistic regression.
\newblock {\em The Annals of Statistics\/}~{\em 48\/}(1), 27–42.

\bibitem[\protect\citeauthoryear{Celentano, Montanari, and Wei}{Celentano
  et~al.}{2020}]{celentano2020lasso}
Celentano, M., A.~Montanari, and Y.~Wei (2020).
\newblock The lasso with general gaussian designs with applications to
  hypothesis testing.
\newblock {\em arXiv preprint arXiv:2007.13716\/}.

\bibitem[\protect\citeauthoryear{Deng, Kammoun, and Thrampoulidis}{Deng
  et~al.}{2019}]{dengmodel}
Deng, Z., A.~Kammoun, and C.~Thrampoulidis (2019).
\newblock A model of double descent for high-dimensional binary linear
  classification.
\newblock {\em Information and Inference: A Journal of the IMA\/}.

\bibitem[\protect\citeauthoryear{Dobriban and Wager}{Dobriban and
  Wager}{2018}]{dobriban2018high}
Dobriban, E. and S.~Wager (2018).
\newblock High-dimensional asymptotics of prediction: Ridge regression and
  classification.
\newblock {\em The Annals of Statistics\/}~{\em 46\/}(1), 247–279.

\bibitem[\protect\citeauthoryear{{Dongning Guo} and {Verdu}}{{Dongning Guo} and
  {Verdu}}{2005}]{Guo}
{Dongning Guo} and S.~{Verdu} (2005).
\newblock Randomly spread cdma: asymptotics via statistical physics.
\newblock {\em IEEE Transactions on Information Theory\/}~{\em 51\/}(6),
  1983--2010.

\bibitem[\protect\citeauthoryear{Fern\'{a}ndez-Delgado, Cernadas, Barro, and
  Amorim}{Fern\'{a}ndez-Delgado et~al.}{2014}]{JMLR:v15:delgado14a}
Fern\'{a}ndez-Delgado, M., E.~Cernadas, S.~Barro, and D.~Amorim (2014).
\newblock Do we need hundreds of classifiers to solve real world classification
  problems?
\newblock {\em Journal of Machine Learning Research\/}~{\em 15}, 3133--3181.

\bibitem[\protect\citeauthoryear{Freund and Schapire}{Freund and
  Schapire}{1997}]{FREUND1997119}
Freund, Y. and R.~E. Schapire (1997).
\newblock A decision-theoretic generalization of on-line learning and an
  application to boosting.
\newblock {\em Journal of Computer and System Sciences\/}~{\em 55\/}(1), 119 --
  139.

\bibitem[\protect\citeauthoryear{Friedman, Hastie, and Tibshirani}{Friedman
  et~al.}{2000}]{friedman2000}
Friedman, J., T.~Hastie, and R.~Tibshirani (2000).
\newblock Additive logistic regression: a statistical view of boosting.
\newblock {\em Annals of Statistics\/}~{\em 28}, 337--407.

\bibitem[\protect\citeauthoryear{Gerace, Loureiro, Krzakala, Mézard, and
  Zdeborová}{Gerace et~al.}{2020}]{gerace2020generalisation}
Gerace, F., B.~Loureiro, F.~Krzakala, M.~Mézard, and L.~Zdeborová (2020).
\newblock Generalisation error in learning with random features and the hidden
  manifold model.
\newblock In {\em International Conference on Machine Learning}, pp.\
  3452–3462. PMLR.

\bibitem[\protect\citeauthoryear{Gerbelot, Abbara, and Krzakala}{Gerbelot
  et~al.}{2020}]{gerbelot2020asymptotic}
Gerbelot, C., A.~Abbara, and F.~Krzakala (2020).
\newblock Asymptotic errors for teacher-student convex generalized linear
  models (or: How to prove kabashima’s replica formula).
\newblock {\em arXiv preprint arXiv:2006.06581\/}.

\bibitem[\protect\citeauthoryear{Hastie, Buja, and Tibshirani}{Hastie
  et~al.}{1995}]{10.2307/2242400}
Hastie, T., A.~Buja, and R.~Tibshirani (1995).
\newblock Penalized discriminant analysis.
\newblock {\em The Annals of Statistics\/}~{\em 23\/}(1), 73--102.

\bibitem[\protect\citeauthoryear{Hastie, Montanari, Rosset, and
  Tibshirani}{Hastie et~al.}{2019}]{hastie2019surprises}
Hastie, T., A.~Montanari, S.~Rosset, and R.~J. Tibshirani (2019).
\newblock Surprises in high-dimensional ridgeless least squares interpolation.
\newblock {\em arXiv preprint arXiv:1903.08560\/}.

\bibitem[\protect\citeauthoryear{Huang}{Huang}{2017}]{Huang17}
Huang, H. (2017).
\newblock Asymptotic behavior of support vector machine for spiked population
  model.
\newblock {\em Journal of Machine Learning Research\/}~{\em 18}, 45:1--45:21.

\bibitem[\protect\citeauthoryear{Huang}{Huang}{2018}]{8450750}
Huang, H. (2018, June).
\newblock Asymptotic behavior of margin-based classification methods.
\newblock In {\em 2018 IEEE Statistical Signal Processing Workshop (SSP)}, pp.\
   463--467.

\bibitem[\protect\citeauthoryear{Huang, Liu, Yuan, and Marron}{Huang
  et~al.}{2015}]{softthreshold}
Huang, H., Y.~Liu, M.~Yuan, and J.~S. Marron (2015).
\newblock Statistical significance of clustering using soft thresholding.
\newblock {\em Journal of Computational and Graphical Statistics\/}~{\em
  24\/}(4), 975--993.

\bibitem[\protect\citeauthoryear{Johnstone}{Johnstone}{2001}]{Johnstone}
Johnstone, I.~M. (2001).
\newblock {On the Distribution of the Largest Eigenvalue in Principal
  Components Analysis}.
\newblock {\em The Annals of Statistics\/}~{\em 29\/}(2), 295--327.

\bibitem[\protect\citeauthoryear{Laloux, Cizeau, Potters, and Bouchaud}{Laloux
  et~al.}{2000}]{doi:10.1142/S0219024900000255}
Laloux, L., P.~Cizeau, M.~Potters, and J.-P. Bouchaud (2000).
\newblock Random matrix theory and financial correlations.
\newblock {\em International Journal of Theoretical and Applied Finance\/}~{\em
  03\/}(03), 391--397.

\bibitem[\protect\citeauthoryear{Lam, Marron, Sun, and Toh}{Lam
  et~al.}{2018}]{Defeng}
Lam, X.~Y., J.~S. Marron, D.~Sun, and K.-C. Toh (2018).
\newblock Fast algorithms for large-scale generalized distance weighted
  discrimination.
\newblock {\em Journal of Computational and Graphical Statistics\/}~{\em
  27\/}(2), 368--379.

\bibitem[\protect\citeauthoryear{Lin, Wahba, Xiang, Gao, Klein, and Klein}{Lin
  et~al.}{2000}]{lin2000}
Lin, X., G.~Wahba, D.~Xiang, F.~Gao, R.~Klein, and B.~Klein (2000).
\newblock Smoothing spline anova models for large data sets with bernoulli
  observations and the randomized gacv.
\newblock {\em The Annals of Statistics\/}~{\em 28\/}(6), 1570--1600.

\bibitem[\protect\citeauthoryear{Liu, Parker, Fan, Perou, and Marron}{Liu
  et~al.}{2009}]{liuxy}
Liu, X., J.~Parker, C.~Fan, C.~M. Perou, and J.~S. Marron (2009).
\newblock {\em Visualization of Cross-Platform Microarray Normalization},
  Chapter~14, pp.\  167--181.
\newblock John Wiley and Sons, Ltd.

\bibitem[\protect\citeauthoryear{Liu, Hayes, Nobel, and Marron}{Liu
  et~al.}{2008}]{liu:sigclust}
Liu, Y., D.~N. Hayes, A.~Nobel, and J.~S. Marron (2008).
\newblock Statistical significance of clustering for high-dimension, low-sample
  size data.
\newblock {\em Journal of the American Statistical Association\/}~{\em
  103\/}(483), 1281--1293.

\bibitem[\protect\citeauthoryear{Liu, Zhang, and Wu}{Liu
  et~al.}{2011}]{liu2011}
Liu, Y., H.~H. Zhang, and Y.~Wu (2011).
\newblock Soft or hard classification? large margin unified machines.
\newblock {\em Journal of the American Statistical Association\/}~{\em 106},
  166--177.

\bibitem[\protect\citeauthoryear{Loureiro, Gerbelot, Cui, Goldt, Krzakala,
  Mézard, and Zdeborová}{Loureiro et~al.}{2021a}]{loureiro2021capturing}
Loureiro, B., C.~Gerbelot, H.~Cui, S.~Goldt, F.~Krzakala, M.~Mézard, and
  L.~Zdeborová (2021a).
\newblock Learning curves of generic features maps for realistic datasets with
  a teacher-student model.
\newblock {\em arXiv preprint arXiv:2102.08127\/}.

\bibitem[\protect\citeauthoryear{Loureiro, Sicuro, Gerbelot, Pacco, Krzakala,
  and Zdeborová}{Loureiro et~al.}{2021b}]{loureiro2021learning}
Loureiro, B., G.~Sicuro, C.~Gerbelot, A.~Pacco, F.~Krzakala, and L.~Zdeborová
  (2021b).
\newblock Learning gaussian mixtures with generalised linear models: Precise
  asymptotics in high-dimensions.
\newblock {\em arXiv preprint arXiv:2106.03791\/}.

\bibitem[\protect\citeauthoryear{Ma}{Ma}{2013}]{ma2013}
Ma, Z. (2013, 04).
\newblock Sparse principal component analysis and iterative thresholding.
\newblock {\em The Annals of Statistics\/}~{\em 41\/}(2), 772--801.

\bibitem[\protect\citeauthoryear{Mai and Couillet}{Mai and
  Couillet}{2018}]{maistatistical}
Mai, X. and R.~Couillet (2018).
\newblock Statistical analysis and improvement of large dimensional svm.
\newblock {\em private communication\/}.

\bibitem[\protect\citeauthoryear{Mai and Liao}{Mai and
  Liao}{2019}]{Mai2019HighDC}
Mai, X. and Z.~Liao (2019).
\newblock High dimensional classification via empirical risk minimization:
  Improvements and optimality.
\newblock {\em ArXiv\/}~{\em abs/1905.13742}.

\bibitem[\protect\citeauthoryear{Mai, Liao, and Couillet}{Mai
  et~al.}{2019}]{mailiao}
Mai, X., Z.~Liao, and R.~Couillet (2019, May).
\newblock A large scale analysis of logistic regression: Asymptotic performance
  and new insights.
\newblock In {\em ICASSP 2019 - 2019 IEEE International Conference on
  Acoustics, Speech and Signal Processing (ICASSP)}, pp.\  3357--3361.

\bibitem[\protect\citeauthoryear{Marcenko and Pastur}{Marcenko and
  Pastur}{1967}]{Pastur}
Marcenko, V.~A. and L.~A. Pastur (1967).
\newblock Distribution of eigenvalues for some sets of random matrices.
\newblock {\em Mathematics of the USSR-Sbornik\/}~{\em 1\/}(4), 457--483.

\bibitem[\protect\citeauthoryear{Marron, Todd, and Ahn}{Marron
  et~al.}{2007}]{Marron2007}
Marron, J.~S., M.~Todd, and J.~Ahn (2007).
\newblock Distance-weighted discrimination.
\newblock {\em Journal of the American Statistical Association\/}~{\em 102},
  1267--1271.

\bibitem[\protect\citeauthoryear{M{\'e}zard and Montanari}{M{\'e}zard and
  Montanari}{2009}]{mezard2009information}
M{\'e}zard, M. and A.~Montanari (2009).
\newblock {\em Information, Physics, and Computation}.
\newblock Oxford Graduate Texts. OUP Oxford.

\bibitem[\protect\citeauthoryear{Mezard, Parisi, and Virasoro}{Mezard
  et~al.}{1987}]{mezard1987spin}
Mezard, M., G.~Parisi, and M.~Virasoro (1987).
\newblock {\em Spin Glass Theory and Beyond: An Introduction to the Replica
  Method and Its Applications}.
\newblock World Scientific Lecture Notes in Physics. World Scientific.

\bibitem[\protect\citeauthoryear{Mignacco, Krzakala, Lu, Urbani, and
  Zdeborova}{Mignacco et~al.}{2020}]{mignacco2020role}
Mignacco, F., F.~Krzakala, Y.~Lu, P.~Urbani, and L.~Zdeborova (2020).
\newblock The role of regularization in classification of high-dimensional
  noisy gaussian mixture.
\newblock In {\em International Conference on Machine Learning}, pp.\
  6874–6883. PMLR.

\bibitem[\protect\citeauthoryear{Montanari, Ruan, Sohn, and Yan}{Montanari
  et~al.}{2019}]{montanari2019generalization}
Montanari, A., F.~Ruan, Y.~Sohn, and J.~Yan (2019).
\newblock The generalization error of maxmargin linear classifiers:
  High-dimensional asymptotics in the overparametrized regime.
\newblock {\em arXiv preprint arXiv:1911.01544\/}.

\bibitem[\protect\citeauthoryear{Qiao, Zhang, Liu, Todd, and Marron}{Qiao
  et~al.}{2010}]{Qiao2011}
Qiao, X., H.~H. Zhang, Y.~Liu, M.~J. Todd, and J.~S. Marron (2010).
\newblock Asymptotic properties of distance-weighted discrimination.
\newblock {\em Journal of the American Statistical Association\/}~{\em
  105\/}(489), 401--414.

\bibitem[\protect\citeauthoryear{Qiao and Zhang}{Qiao and Zhang}{2015}]{qiao15}
Qiao, X. and L.~Zhang (2015).
\newblock Flexible high-dimensional classification machines and their
  asymptotic properties.
\newblock {\em Journal of Machine Learning Research\/}~{\em 16}, 1547--1572.

\bibitem[\protect\citeauthoryear{Sear and Cuesta}{Sear and
  Cuesta}{2003}]{PhysRevLett.91.245701}
Sear, R.~P. and J.~A. Cuesta (2003).
\newblock Instabilities in complex mixtures with a large number of components.
\newblock {\em Phys. Rev. Lett.\/}~{\em 91}, 245701.

\bibitem[\protect\citeauthoryear{Shen, Tseng, Zhang, and Wong}{Shen
  et~al.}{2003}]{shen2003}
Shen, X., G.~C. Tseng, X.~Zhang, and W.~H. Wong (2003).
\newblock On $\psi$-learning.
\newblock {\em Journal of the American Statistical Association\/}~{\em
  98\/}(463), 724--734.

\bibitem[\protect\citeauthoryear{{Sifaou}, {Kammoun}, and {Alouini}}{{Sifaou}
  et~al.}{2019}]{9022461}
{Sifaou}, H., A.~{Kammoun}, and M.~{Alouini} (2019).
\newblock Phase transition in the hard-margin support vector machines.
\newblock In {\em 2019 IEEE 8th International Workshop on Computational
  Advances in Multi-Sensor Adaptive Processing (CAMSAP)}, pp.\  415--419.

\bibitem[\protect\citeauthoryear{Tanaka}{Tanaka}{2002}]{Tanaka}
Tanaka, T. (2002).
\newblock A statistical-mechanics approach to large-system analysis of cdma
  multiuser detectors.
\newblock {\em Information Theory, IEEE Transactions on\/}~{\em 48\/}(11),
  2888--2910.

\bibitem[\protect\citeauthoryear{TCGA}{TCGA}{2010}]{tcga2010}
TCGA (2010).
\newblock The cancer genome atlas research network.
\newblock {\em
  http://cancergenome.nih.gov/wwd/pilot\_program/research\_network/cgcc.asp\/}.

\bibitem[\protect\citeauthoryear{Telatar}{Telatar}{1999}]{citeulike:2714811}
Telatar, E. (1999, November).
\newblock Capacity of multi-antenna {Gaussian} channels.
\newblock {\em Eur. Trans. Telecomm. ETT\/}~{\em 10\/}(6), 585--596.

\bibitem[\protect\citeauthoryear{Vapnik}{Vapnik}{1995}]{Vapnik95}
Vapnik, V.~N. (1995).
\newblock {\em The Nature of Statistical Learning Theory}.
\newblock New York, NY: Springer.

\bibitem[\protect\citeauthoryear{Wang and Zou}{Wang and Zou}{2016}]{zou2}
Wang, B. and H.~Zou (2016).
\newblock Sparse distance weighted discrimination.
\newblock {\em Journal of Computational and Graphical Statistics\/}~{\em
  25\/}(3), 826--838.

\bibitem[\protect\citeauthoryear{Wang and Zou}{Wang and Zou}{2018}]{zou1}
Wang, B. and H.~Zou (2018).
\newblock Another look at distance-weighted discrimination.
\newblock {\em Journal of the Royal Statistical Society: Series B (Statistical
  Methodology)\/}~{\em 80\/}(1), 177--198.

\bibitem[\protect\citeauthoryear{Wang and Thrampoulidis}{Wang and
  Thrampoulidis}{2021}]{wang2021binary}
Wang, K. and C.~Thrampoulidis (2021).
\newblock Binary classification of gaussian mixtures: Abundance of support
  vectors, benign overfitting and regularization.

\bibitem[\protect\citeauthoryear{Wu and Verdu}{Wu and Verdu}{2012}]{Verdu}
Wu, Y. and S.~Verdu (2012, October).
\newblock Optimal phase transitions in compressed sensing.
\newblock {\em IEEE Transactions on Information Theory\/}~{\em 58\/}(10),
  6241–6263.

\bibitem[\protect\citeauthoryear{Zhu and Hastie}{Zhu and
  Hastie}{2005}]{doi:10.1198/106186005X25619}
Zhu, J. and T.~Hastie (2005).
\newblock Kernel logistic regression and the import vector machine.
\newblock {\em Journal of Computational and Graphical Statistics\/}~{\em
  14\/}(1), 185--205.

\end{thebibliography}
\end{document}